\pdfoutput=1

\documentclass[11pt]{article}

\usepackage[final]{acl}

\usepackage{times}
\usepackage{latexsym}
\usepackage{adjustbox}
\usepackage{placeins}

\usepackage[T1]{fontenc}

\usepackage[utf8]{inputenc}

\usepackage{microtype}

\usepackage{inconsolata}

\usepackage{graphicx}

\usepackage{amsmath}

\usepackage{tcolorbox}
\usepackage{booktabs}
\tcbuselibrary{skins}    
\usepackage{multirow}    
\usepackage{tabularx}
\usepackage{colortbl}
\usepackage{longtable}
\title{Multi\texorpdfstring{$^2$}{2}: Multi-Agent Test-Time Scalable Framework for \\ Multi-Document Processing}

\author{
Juntai Cao$^{1*}$,  Xiang Zhang$^{1}$\thanks{Equal contribution.},  Raymond Li$^{1}$, Chuyuan Li$^{1}$, Chenyu You$^{2}$, Shafiq Joty$^{3}$, Giuseppe Carenini$^{1}$ \\ 
$^{1}$ University of British Columbia \\
$^{2}$ Stony Brook University \\
$^{3}$ Salesforce Research \\ 
\texttt{\{jtcao7, raymondl,
chuyuan.li, carenini\}@cs.ubc.ca} \\ \texttt{xzhang23@ualberta.ca}, \ \texttt{cyou@cs.stonybrook.edu}, \ \texttt{sjoty@salesforce.com}}

\begin{document}
\maketitle
\begin{abstract}

Recent advances in test-time scaling have shown promising results in improving Large Language Model (LLM) performance through strategic computation allocation during inference. While this approach has demonstrated strong improvements in logical and mathematical reasoning tasks, its application to natural language generation (NLG), particularly summarization, remains unexplored.
Multi-Document Summarization (MDS), a fundamental task in NLG, presents unique challenges by requiring models to extract and synthesize essential information across multiple lengthy documents. Unlike reasoning tasks, MDS demands a more nuanced approach to prompt design and ensemble methods, as no single "best" prompt can satisfy diverse summarization requirements.
We propose a novel framework leveraging test-time scaling for MDS. Our approach employs prompt ensemble techniques to generate multiple candidate summaries using various prompts, then combines them with an aggregator to produce a refined summary. To evaluate our method effectively, we also introduce two new LLM-based metrics: the Consistency-Aware Preference (CAP) score and LLM Atom-Content-Unit (LLM-ACU) score, which assess summary quality while addressing the positional bias inherent in traditional automatic evaluation.
Our extensive experiments demonstrate that this framework significantly enhances summary quality while also revealing the practical scaling boundaries to MDS tasks. Our code, results, and evaluation module will be released at \href{https://anonymous.4open.science/r/Multi2-F7E7/README.md}{anonymous Github}.
\end{abstract}

\section{Introduction}

Test-time scaling (or inference-time scaling) has emerged as a promising approach for enhancing LLM's performance beyond traditional architectural or data improvements \cite{learningtoreason2024}. While earlier work focused on relationships between models' capabilities, size, and training resources, recent research demonstrates that strategic compute allocation during inference can yield substantial performance gains. For instance, studies show that increased inference computation produces better results than equivalent investments in pretraining 
\citep{snell2024scalingllmtesttimecompute,agarwal2024manyshotincontextlearning, muennighoff2025s1simpletesttimescaling}.
\begin{figure}[!t]
    \centering
    \includegraphics[width=0.86\linewidth]{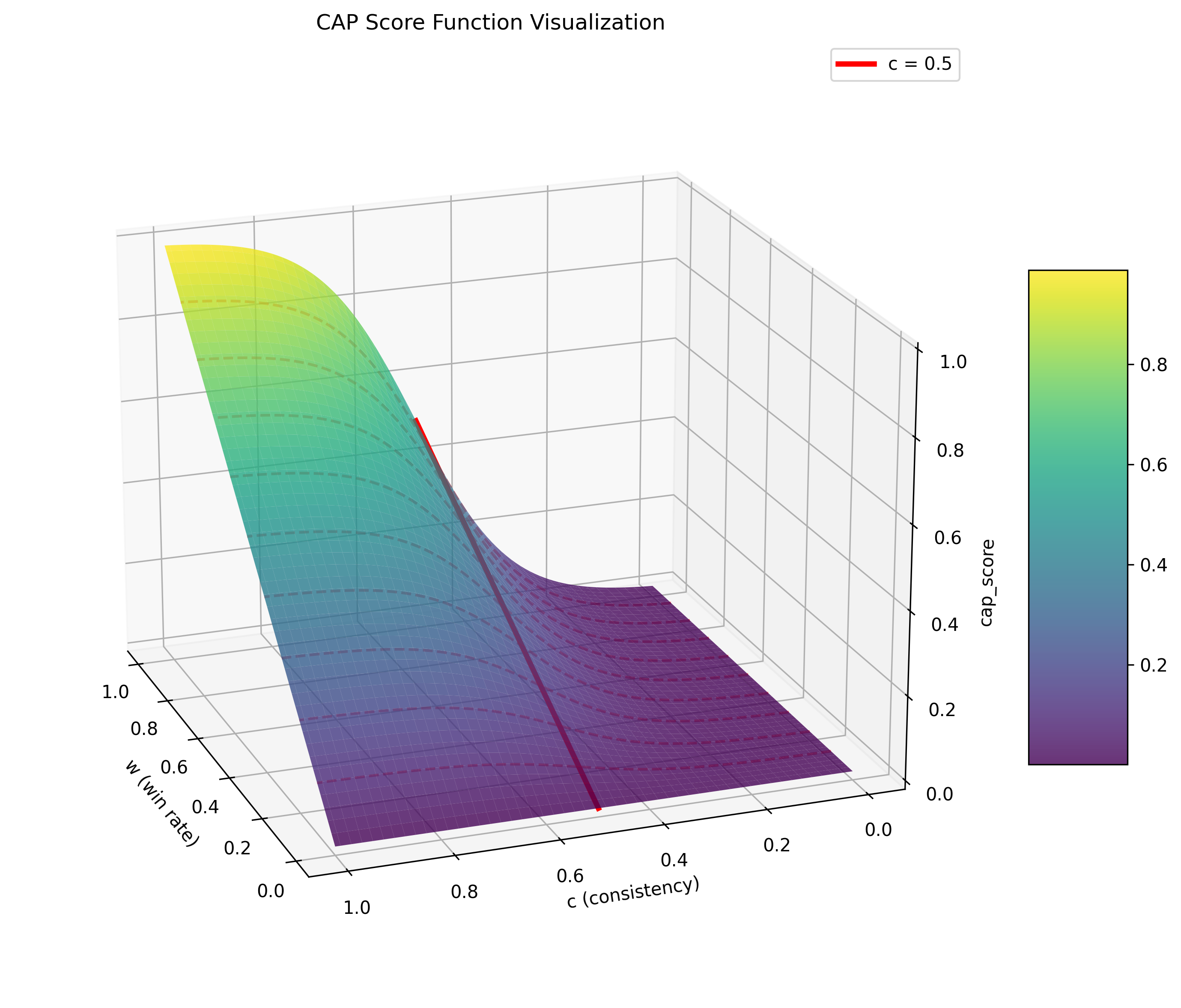}
    \caption{
    Visualization of our proposed Consistency-aware Preference (CAP) Score for text generation task. Applying LLMs' strong language understanding ability, CAP assign higher score to summary which 
    \textit{consistently} gets ranked higher by the LLM.}
    \label{fig:cap_viz}
\end{figure}

Research on test-time scaling has largely centered on logical and math reasoning tasks, leaving traditional natural language generation (NLG) tasks relatively unexplored. 
This gap is particularly notable in summarization, a domain where LLMs have already demonstrated significant advances, generating summaries competitive with human performance \citep{xiao-etal-2024-personalized,zhang-etal-2024-benchmarking,pu2023summarizationalmostdead}.
Beyond text generation, LLMs have also been proven effective as judges when guided by well-designed evaluation protocols \citep{liu-etal-2024-benchmarking,liu2024reifereevaluatinginstructionfollowingevaluation}. 
Recent expansions in context window sizes have created new opportunities to study scaling effects on length-constrained tasks like summarization \citep{liu2022character}. However, LLMs still struggle with key challenges including hallucination, incomplete coverage, language inconsistency, and verbosity \citep{zhang2024cross,liu-etal-2024-benchmarking,zhang2023bridging,belem2024singlemultillmshallucinate,zhang-etal-2023-dont}.

In this paper, we aim to examine LLMs' summarization capabilities and their scaling properties by focusing on the multi-document summarization (MDS) task. 
MDS requires synthesizing and linking information across lengthy documents, handling information redundancy, maintaining factual consistency, and generating coherent and concise summaries while preserving key details. 
In addition, MDS demands effective reasoning to determine relevance and priority among diverse pieces of information.
These characteristics make MDS particularly time- and labor-intensive \citep{vanveen2024clinical}. 
To tackle these challenges, we propose a multi-agent approach that leverages prompt ensemble to scale summarization at test time. While traditional prompt ensemble methods exist—such as (a) applying different sampling strategies to a single prompt \citep{li2023makinglargelanguagemodels}, or (b) varying few-shot examples within prompts \citep{arora2022askanythingsimplestrategy}—their direct application to summarization presents notable limitations. The first approach merely explores variations in the output space, while the second heavily relies on example-based learning, which is better suited for reasoning tasks \citep{zhang2024supervisedchainthought}.
Furthermore, summarization differs fundamentally from reasoning tasks~\cite{zhang2024counting}, where specific prompts like ``\textit{Let's think step by step}'' \citep{kojima2022large} can effectively guide models through predetermined reasoning patterns~\cite{zhang2024autoregressive+}. In contrast, no single "optimal" prompt exists for generating summaries that satisfy diverse requirements. Given these distinctions, summarization demands a more sophisticated approach to prompt ensemble techniques. The theory behind prompt-space ensemble in explained in Appendix~\ref{app:theory}.

\begin{figure*}
    \centering
    \includegraphics[width=\linewidth]{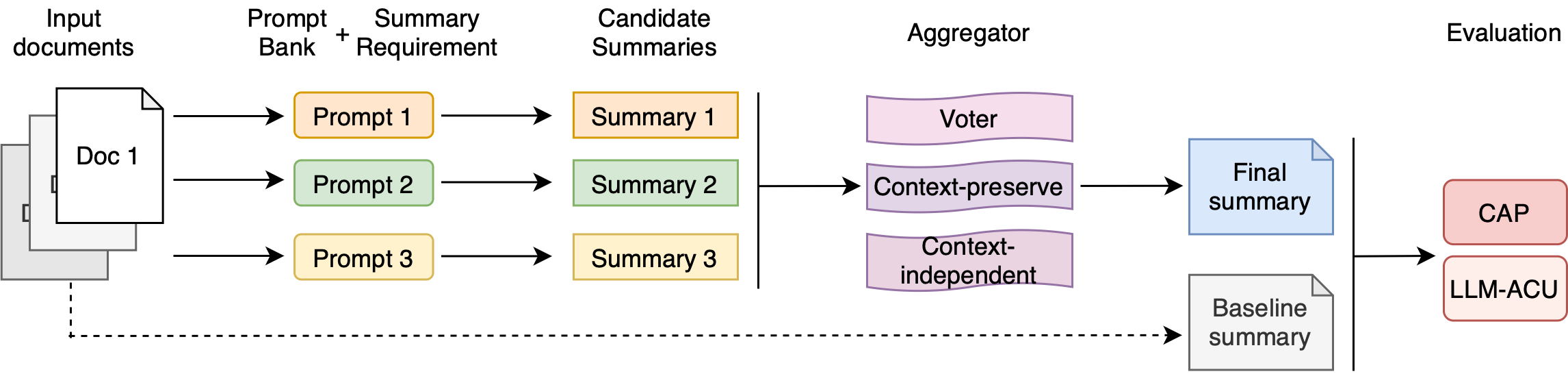}
    \caption{
    Overview of Multi$^2$ summarization inference-time scaling framework. Documents are first summarized by independent LLM agents, each guided by a different prompt from a curated prompt bank and constrained by user requirements. 
    The resulting 
    summaries are then processed by an aggregator (Voter, Context-Preserving Summarizer, or Context-Independent Summarizer)
    to generate the final consolidated summary.
    }
    \label{fig:framework}
\end{figure*}

Therefore,
we propose Multi$^2$ framework (Fig. \ref{fig:framework}) to 
address this challenge by generating multiple summaries through diverse prompts while maintaining consistent requirements. 
We then employ an aggregation strategy to construct a comprehensive final summary that leverages the strengths of each summary candidate.
While increased inference-time computation generally improves performance, recent studies have also identified an \textit{inverse scaling} phenomenon, where excessive computation can paradoxically degrade performance \citep{gao2022scalinglawsrewardmodel, stroebl2024inferencescalingflawslimits}. 
We also investigate this phenomenon by systematically varying the number of samples and examining its boundaries. 

Another challenge in MDS is the reliability of automatic evaluation metrics. Traditional metrics like ROUGE \citep{lin2004rouge} have proven insufficient for capturing summary quality, while more recent LLM-based metrics—such as Auto-ACU \citep{liu-etal-2023-towards-interpretable}, LLMCompare \citep{liu-etal-2024-benchmarking}, and LLMRank \citep{liu2024reifereevaluatinginstructionfollowingevaluation}—show limitations, including constraints in contextual understanding for smaller models  and persistent positional biases \citep{wang-etal-2024-large-language-models-fair}.  
We specifically highlight \textbf{positional bias}, where LLMs tend to favor summaries appearing in a particular position (first or second in a pairwise comparison), leading to inconsistencies in evaluation, particularly during test-time scaling.  
To improve evaluation consistency, we propose two novel metrics: Consistency-Aware Preference (CAP) score and LLM Atom-Content-Unit (LLM-ACU) score. As shown in our experiments, these metrics leverage LLMs' contextual understanding while incorporating mechanisms to mitigate positional bias, ensuring more reliable and robust summary assessment.

In summary, (1) We present the first comprehensive investigation of test-time scaling laws in text summarization, extending the analysis beyond traditionally explored reasoning tasks; (2) We introduce a new framework Multi$^2$ that enhances summarization performance through prompt ensemble at test time; (3) We enhance two existing evaluation protocols for summarization through strategic modifications and incorporating LLMs, improving quantitative assessment of summary quality and advancing automatic evaluation methodologies for summarization tasks.

\section{Formulation of Prompt Ensemble}

Let \( x \) denote the input text and \(\mathcal{P} = \{ p_1, p_2, \ldots, p_N \}\) be a collection of prompts designed to elicit different aspects of information from the underlying language model. For each prompt \( p_i \in \mathcal{P} \), the model produces an output \( y_i \) according to a generation function \( f \):
\[
y_i = f(x, p_i).
\]
The intuition behind this methodology is that different prompts \( p_i \) induce the model to focus on distinct features or details in the input \( x \), thereby generating complementary outputs.

To combine these outputs, we define an aggregation function \( g: \mathcal{Y}^N \to \mathcal{Y} \) that fuses the individual outputs \(\{ y_1, y_2, \ldots, y_N \}\) into a final output \( y \):
\[
y = g(y_1, y_2, \ldots, y_N).
\]
The aggregation function \(g\) can take various forms depending on the specific application, with weighted averaging and majority voting being common implementations. For our MDS task, we implement three distinct formulations of \(g\): content-independent summarization, content-preserving summarization, and voting-based aggregation. The overall system can therefore be formalized as:
\[
y = g\big(f(x, p_1), f(x, p_2), \ldots, f(x, p_N)\big).
\]

This formulation ensures that the final generated text \( y \) benefits from the diverse perspectives provided by the prompt ensemble. Empirical results indicate that the ensemble method consistently outperforms individual prompt-based generations, as it effectively mitigates the shortcomings of any single prompt by incorporating a broader range of contextual insights from the input \( x \). A more detailed explanation for prompt space theory are attached in Appendix~\ref{app:theory}.

\section{Multi\texorpdfstring{$^2$}{2} Framework}

\subsection{Multi-Agent Summarization}
Our Multi$^2$ test-time scaling framework for MDS is illustrated in Figure~\ref{fig:framework}. The framework operates in two main stages: candidate generation and summary aggregation. In the first stage, input documents are processed by multiple independent LLM agents using randomly selected prompts from a curated prompt bank, simulating real-world summarization scenarios. The generated candidate summaries, along with the original requirements, are then passed to the aggregator module.
The aggregator module implements three distinct approaches: vote, context-preserving summarizer (CPS), and context-independent summarizer (CIS). 

The \textbf{vote} agent evaluates all candidate summaries against the original input documents and provides a detailed explanation before selecting the best summary. To mitigate positional bias (Appendix~\ref{app:posbias}), we explicitly require the agent to complete its reasoning before indicating its final selection, ensuring the choice is constrained by the documented rationale. 
Instead of selecting the best candidate summary, CPS and CIS aggregate the candidate summaries into a final summary. 
The \textbf{CPS} agent generates a refined summary by consulting both the original documents and the candidate summaries, aiming for completeness and conciseness. 
In contrast, the \textbf{CIS} agent focuses solely on the candidate summaries without access to the original documents, producing a consolidated summary through reference-based synthesis. We attached our prompts for aggregation agents in Appendix~\ref{app:ensembleprompt}.

\subsection{Automatic Evaluation}

\subsubsection{Positional Bias and Motivation} 
\label{sec:metricmotiv}
Recent approaches to automatic evaluation have increasingly leveraged LLMs, either through comparative (pairwise) assessment or direct scoring mechanisms. However, both approaches face  challenges. Comparative methods struggle with positional bias, an inherent limitation of LLM judges. While previous research \cite{liu2024reifereevaluatinginstructionfollowingevaluation} suggested that advanced models (like \texttt{gpt-4o}) might mitigate this issue, our experiments in Appendix~\ref{app:posbias} demonstrate that LLM evaluations remain extremely susceptible to position-dependent variations, especially on contextual tasks like MDS. 
Direct scoring approaches face different challenges: defining clear scoring guidelines could be difficult, and ensuring consistent application of grading rubrics across different generations remains challenging. Moreover, the complexity of nuanced scoring—a task challenging even for human evaluators who struggle more with five-point Likert scales than binary preferences—makes it particularly difficult for LLMs to provide reliable quantitative assessments. The pairwise comparative setup offers utility to practitioners (e.g., evaluation for A/B testing) while eliciting evaluations better aligned with humans judgment from automatic evaluators~\citep{wang2023large,liu2024aligning}.

To address these limitations and enable reliable large-scale evaluation of generated summaries, we propose two novel metrics 
Consistency-aware Preference (CAP) score and LLM-ACU score. These metrics are specifically designed to mitigate positional bias, while providing repeatable quantitative measurements for systematic comparison of summary quality.

\subsubsection{Consistency-Aware Preference Score}
\begin{figure}
    \centering
    \includegraphics[width=\columnwidth]{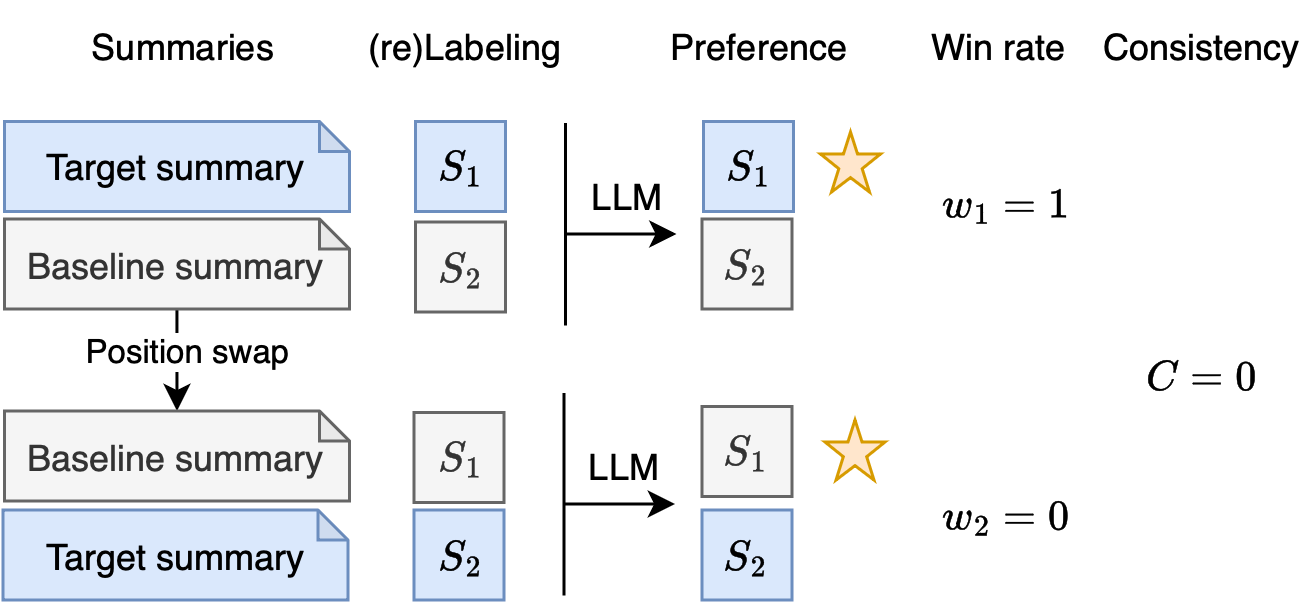}
    \caption{Two-step calculation of CAP using LLM. In this example, LLM prefers the first summary in both step 1 and step 2, resulting in inconsistent evaluation.}
    \label{fig:cap1}
\end{figure}

We develop the Consistency-Aware Preference (CAP) score as an enhancement to the LLMCompare \citep{liu-etal-2024-benchmarking} method for quantitatively evaluating preference rates of summaries compared to a baseline. 
LLMCompare employs an LLM judge to evaluate two summaries against the source documents, determining which is superior (1 or 2) or if they are equivalent (tie). 
To address the positional bias (detailed in Section~\ref{sec:metricmotiv}), we implement a metric with two-phase comparison process. 
First, we use an LLM as judge to obtain preferences with summaries (target and baseline) in their original positions. 
Then, we swap the positions of the two summaries and obtain a second set of preferences, relabeling them based on their new positions to eliminate labeling bias. 
From this two-step comparison, we compute the \textit{win rates} ($w_1$, $w_2$) of the target summarization method against the baseline in 
each step,
and the \textit{consistency rate} ($C$) of predictions across both orderings (Figure~\ref{fig:cap1}).

Importantly,
when evaluating consistency, if either comparison (i.e., before or after the swapping) results in a tie, we consider it consistent with any outcome in the other comparison to avoid over-penalizing borderline cases.
The final CAP score is computed as follows:
\begin{equation}
\text{CAP} = W_{\text{pref}}\frac{1}{1+\exp^{-k(C-0.5)}},
\end{equation}
where $W_{\text{pref}}$ refers to \textit{preference rate} calculated from \textit{win rates} ($w_1$ and $w_2$); $C$ refers to \textit{consistency score}; 
$k$ controls sensitivity to consistency variations (default to 10 according to the our experiments on a validation set).
In practice, the preference weight $W_{\text{pref}}$ can be determined using either max-pooling or averaging:
\begin{align}
    W_{\text{pref}}^{\text{max}} &= \max(w_1,w_2) \\
    W_{\text{pref}}^{\text{avg}} &= \frac{(w_1,w_2)}{2}
\end{align} 
Figure~\ref{fig:cap_viz} illustrates the distribution of CAP score across different preference weights $W$ and consistency values $C$. 

The CAP score integrates both \textit{preference rate} and \textit{consistency} to ensure robust evaluation. A high CAP score requires both factors to be high, indicating consistent preference for the same summary. When model predictions remain stable, the CAP score correlates directly with the \textit{preference rate}. However, inconsistent predictions yield low CAP scores regardless of preference outcomes, as the metric deliberately penalizes unreliable evaluations.

A near-zero CAP score can result from either (a) summaries that consistently underperform the baseline, or (b) unreliable evaluations due to positional bias. Our framework effectively distinguishes between these scenarios. For instance, if evaluations consistently favor Position 1 regardless of content, the win rate might reach 100\%, but the consistency score would approach 0, yielding a very low CAP score (~0.06)—correctly identifying unreliable evaluation. Conversely, with high consistency, the same win rate produces a CAP score near 1, indicating clear, reliable preference.

By design, CAP scores above 0.5 indicate performance equal to or better than baseline, while lower scores signal inferior performance or evaluation inconsistency. CAP deliberately employs a conservative approach to ambiguous cases, assigning low scores when no clear winner emerges due to tied quality or inconsistent judgments. This design choice prioritizes robustness and interpretability over sensitivity, treating both ``tie with baseline" and ``worse than baseline" scenarios similarly—both indicate failure to establish consistent advantage.

\subsubsection{LLM-ACU Score}
Inspired by the Atomic Content Unit (ACU) score \citep{liu-etal-2023-revisiting,liu-etal-2023-towards-interpretable}, we propose an LLM-based ACU metric to quantitatively measure the completeness of summaries. The process consists of two phases. First, using few-shot prompting, we guide an LLM to extract ACUs from reference summaries. These ACUs are designed to capture essential factual units that are independently interpretable without
references. In the evaluation phase, we present the extracted ACUs alongside the model generated summary and ask an LLM to determine which of the ACUs are entailed in the generated summary. The final score $f$ for a set of summaries $S$ and their corresponding ACU sets $\mathcal{A}$ is computed as the average unnormalized ACU score:
\begin{equation}
f(S,\mathcal{A}) = \frac{1}{|S|}\sum_{s\in S}\frac{e_s}{|\mathcal{A}_s|},
\end{equation}
where $e_s$ represents the number of ACUs in the system output that are entailed by the gold standard ACUs $\mathcal{A}_s$ determined by the LLM.
We use \texttt{gpt-4o} for both ACU extraction\footnote{Our extracted ACUs for MultiNews and OpenASP datasets will be released.} and entailment verification.

Recent work suggests that fine-tuning primarily enables format adaptation rather than information acquisition in language models \citep{allenzhu2024physicslanguagemodels31}. Therefore, we do not finetune models for extracting ACUs and checking entailment, but instead leverage the advanced language understanding capabilities of LLMs directly for both steps.

\begin{table*}[t]
   \centering
   \setlength{\tabcolsep}{6pt}  
   \begin{tabular}{c cc cccc cccc cc}
   \toprule
   & \multicolumn{2}{c}{\textbf{Baseline}} & \multicolumn{4}{c}{\textbf{gpt-4o}} & \multicolumn{4}{c}{\textbf{gpt-4o-mini}} & \multicolumn{2}{c}{} \\
   \cmidrule(lr){2-3} \cmidrule(lr){4-7} \cmidrule(lr){8-11} \cmidrule(lr){12-13}
   \multirow{2}{*}{\textbf{\# Samples}} & 
   \multirow{2}{*}{\textbf{Max}} & \multirow{2}{*}{\textbf{Avg}} & 
   \multicolumn{2}{c}{\textbf{CIS}} & \multicolumn{2}{c}{\textbf{CPS}} & 
   \multicolumn{2}{c}{\textbf{CIS}} & \multicolumn{2}{c}{\textbf{CPS}} & 
   \multicolumn{2}{c}{\textbf{Vote}} \\
   \cmidrule(lr){4-5} \cmidrule(lr){6-7} \cmidrule(lr){8-9} \cmidrule(lr){10-11} \cmidrule(lr){12-13}
   & & & \textbf{Max} & \textbf{Avg} & \textbf{Max} & \textbf{Avg} & 
   \textbf{Max} & \textbf{Avg} & \textbf{Max} & \textbf{Avg} & \textbf{Max} & \textbf{Avg} \\
   \midrule
   2 & \multirow{5}{*}{0.25} & \multirow{5}{*}{0.15} & 0.69 & 0.51 & 0.82 & 0.62 & 0.67 & 0.49 & \textbf{0.80} & \textbf{0.61} & \textbf{0.37} & \textbf{0.23} \\
   3 & & & 0.73 & 0.55 & 0.79 & 0.62 & 0.72 & 0.53 & 0.72 & 0.54 & 0.27 & 0.16 \\
   4 & & & 0.68 & 0.50 & 0.82 & 0.64 & 0.73 & 0.55 & \textbf{0.80} & 0.60 & 0.27 & 0.16 \\
   5 & & & 0.71 & 0.52 & \textbf{0.85} & \textbf{0.69} & \textbf{0.81} & \textbf{0.62} & 0.78 & 0.60 & 0.28 & 0.17 \\
   6 & & & \textbf{0.79} & \textbf{0.60} & 0.81 & 0.63 & 0.77 & 0.57 & 0.77 & 0.60 & \textbf{0.37} & \textbf{0.23} \\
   \bottomrule
   \end{tabular}
   \caption{
    CAP scores on Multinews dataset using \texttt{gpt-4o} and \texttt{gpt-4o-mini} models with context-independent summarizer (CIS) and context-preserving summarizer (CPS). The aggregator using Vote is model-invariant.
   We report CAP with max-pooled (``Max'') and average (``Avg'') preference scores ($W_\text{pref}$).
   Baseline shows both max-pooled and average CAP across all samples. 
   Best scores per column are shown in \textbf{bold}.
   }
   \label{tab:cap_multinews}
\end{table*}

\section{Experiment Setup}

\paragraph{Datasets.} We evaluate our framework on two datasets: MultiNews \citep{fabbri-etal-2019-multi} for general-purpose summarization and OpenASP \citep{amar-etal-2023-openasp} for aspect-based summarization. These datasets represent distinct summarization challenges, with MultiNews focusing on general-purpose news article consolidation and OpenASP targeting aspect-specific information extraction and synthesis. For a balanced comparison, we conduct our experiments on the test sets of both datasets. For MultiNews, we select the first 600 entries from its test set to match the size of OpenASP's test set.

\paragraph{Models.}  To investigate scaling properties and leverage extended context windows, we evaluate our framework using two state-of-the-art models of different scales: \texttt{gpt-4o} and \texttt{gpt-4o-mini}. These models enable us to analyze how performance scales with model size while maintaining consistent architectural characteristics.

\paragraph{Prompt Bank.} We adapt the prompt collection from \citet{lior2024seamstochasticbenchmarkmultidocument} to explore the prompt space. While some prompts in their work were originally designed for extractive summarization, we modified them for abstractive summary generation while preserving their core instructional elements. The prompts are shown in Appendix~\ref{app:summprompt}.

\paragraph{Implementation Details.}
We establish our baseline using summaries generated by \texttt{gpt-4o} with a single prompt randomly selected from our prompt bank using a fixed random seed. We scale summarization by applying different aggregation methods to the generated summaries.
For voting-based aggregation, we exclusively use \texttt{gpt-4o}, since this method operates independently of the generator model and focuses on the well-defined task of selecting the optimal summary from available candidates, rather than producing new text. In contrast, generative aggregation methods synthesize entirely new summaries.
To ensure experimental rigor, we execute each configuration twice with the default temperature setting (0.8) and report averaged results. Our experimental design focuses on two primary variables: (1) inference model size and (2) scaling factor, determined by the number of ensembled samples.

\paragraph{Evaluation Protocols.}
Our experimental evaluation employs multiple complementary metrics: ROUGE Score \citep{lin2004rouge} serves as the traditional lexical overlap measure, while CAP score quantifies user preference compared to the baseline system, and LLM-ACU score assesses information coverage. We employ \texttt{gpt-4o} as our universal evaluator due to its advanced capabilities. A small-scale human evaluation (Appendix~\ref{app:humaneval}) is also performed to corroborate our quantitative results.

\begin{table*}[t]
   \centering
   \setlength{\tabcolsep}{6pt}  
   \begin{tabular}{c cc cccc cccc cc}
   \toprule
   & \multicolumn{2}{c}{\textbf{Baseline}} & \multicolumn{4}{c}{\textbf{gpt-4o}} & \multicolumn{4}{c}{\textbf{gpt-4o-mini}} & \multicolumn{2}{c}{} \\
   \cmidrule(lr){2-3} \cmidrule(lr){4-7} \cmidrule(lr){8-11} \cmidrule(lr){12-13}
   \multirow{2}{*}{\textbf{\# Samples}} & 
   \multirow{2}{*}{\textbf{Max}} & \multirow{2}{*}{\textbf{Avg}} & 
   \multicolumn{2}{c}{\textbf{CIS}} & \multicolumn{2}{c}{\textbf{CPS}} & 
   \multicolumn{2}{c}{\textbf{CIS}} & \multicolumn{2}{c}{\textbf{CPS}} & 
   \multicolumn{2}{c}{\textbf{Vote}} \\
   \cmidrule(lr){4-5} \cmidrule(lr){6-7} \cmidrule(lr){8-9} \cmidrule(lr){10-11} \cmidrule(lr){12-13}
   & & & \textbf{Max} & \textbf{Avg} & \textbf{Max} & \textbf{Avg} & 
   \textbf{Max} & \textbf{Avg} & \textbf{Max} & \textbf{Avg} & \textbf{Max} & \textbf{Avg} \\
   \midrule
   2 & \multirow{5}{*}{0.51} & \multirow{5}{*}{0.36} & 0.63 & 0.50 & 0.70 & 0.55 & 0.73 & 0.57 & 0.79 & 0.63 & 0.61 & 0.45 \\
   3 & & & 0.72 & 0.57 & 0.76 & 0.59 & 0.75 & 0.60 & 0.83 & 0.69 & 0.64 & 0.48 \\
   4 & & & 0.72 & 0.55 & 0.74 & 0.59 & 0.77 & 0.62 & 0.83 & 0.71 & \textbf{0.66} & \textbf{0.51} \\
   5 & & & \textbf{0.74} & 0.59 & 0.76 & \textbf{0.61} & \textbf{0.82} & \textbf{0.67} & \textbf{0.86} & \textbf{0.72} & 0.64 & 0.48 \\
   6 & & & \textbf{0.74} & \textbf{0.60} & \textbf{0.77} & 0.60 & 0.81 & 0.66 & 0.85 & \textbf{0.72} & 0.56 & 0.42 \\
   \bottomrule
   \end{tabular}
   \caption{
   CAP scores on OpenASP dataset under the same settings as in Table~\ref{tab:cap_multinews}.
   }
   \label{tab:cap_openasp}
\end{table*}

\section{Main Results}
Our experimental results are presented in Tables~\ref{tab:cap_multinews} and~\ref{tab:cap_openasp} for preference metric (CAP scores), Tables~\ref{tab:acu_multinews},~\ref{tab:acu_openasp}, and~\ref{tab:rouge_combined} for completeness metrics (LLM-ACU and ROUGE scores), across both MultiNews and OpenASP datasets.
Detailed analyses examining the relationship between summary length and quality are provided in Appendix~\ref{app:lenimpact}.

\subsection{Effectiveness of Test-Time Scaling and Metrics Alignment}
\begin{figure}[htbp]  
    \centering
    \includegraphics[width=0.8\linewidth]{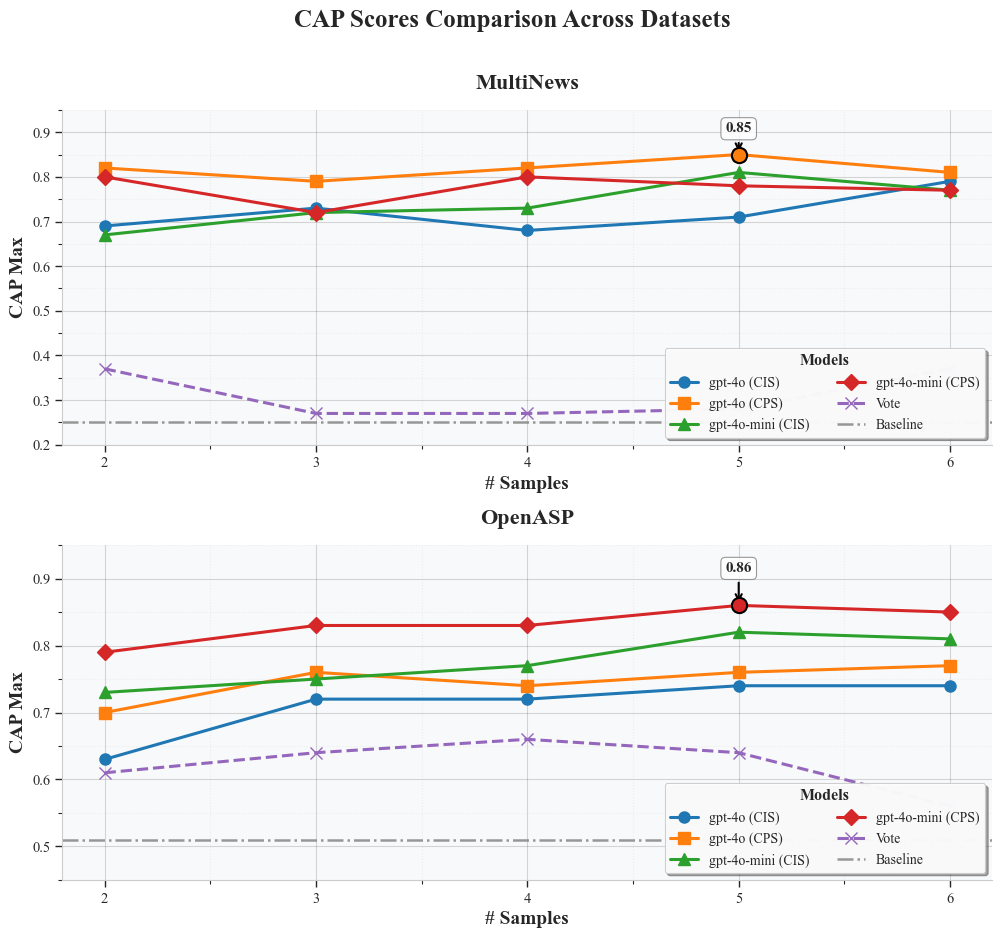}
    \caption{Scaling boundaries based on preference (CAP) scores. For MultiNews, CPS aggregation exhibits inverse scaling effects after 5 samples, yet through ensemble scaling, \texttt{gpt-4o-mini} achieves performance comparable to \texttt{gpt-4o}. In the OpenASP dataset, performance for both models plateaus after 5 samples.}
    \label{fig:capscale}
\end{figure}
Our experiments demonstrate significant improvements through test-time scaling across both preference and completeness metrics.
On MultiNews, starting from a low preference baseline, all scaling methods show substantial gains in overall quality. 
For LLM-ACU score specifically, CPS aggregator achieves the strongest performance in information coverage, with \texttt{gpt-4o-mini} showing substantial gains from a baseline of 47.13 to 54.64 with 6 samples.
Similarly for OpenASP, despite beginning from a stronger preference baseline, scaling with prompt ensemble still provides notable improvements in overall quality.
The LLM-ACU score show comparable trends, with CPS improving \texttt{gpt-4o-mini}'s coverage from 42.35 to 47.82 using 5 samples. These results consistently demonstrate that scaling at test time can effectively enhance both summarization quality and information coverage across different datasets.

Furthermore, Table~\ref{tab:rouge_combined} demonstrates that ROUGE scores consistently improve as the number of ensembled samples increases across both datasets. This trend not only reinforces the effectiveness of our test-time scaling approach from the perspective of traditional metrics, but also validates that our new metrics—CAP and LLM-ACU score—align well with the established ROUGE evaluation framework.

Analysis of the results reveals two key patterns. First, CPS consistently outperforms both CIS and voting approaches across all experimental conditions, suggesting that \textbf{access to source documents during ensemble is crucial for maintaining comprehensive coverage and generating more preferred summaries}. Second, completeness improvements are more pronounced on MultiNews compared to OpenASP, indicating that \textbf{general-purpose summarization may benefit more from diverse prompt sampling for information capture}. 

\begin{table}[t!]
   \centering
   \adjustbox{max width=0.5\textwidth}{%
      \setlength{\tabcolsep}{3pt}  
      \begin{tabular}{c c cc cc c}
         \toprule
         \multicolumn{2}{c}{\textbf{LLM-ACU (MultiNews)}} & \multicolumn{2}{c}{\textbf{gpt-4o}} & \multicolumn{2}{c}{\textbf{gpt-4o-mini}} & \\
         \cmidrule(lr){3-4} \cmidrule(lr){5-6}
         \textbf{\# Samples} & \textbf{Baseline} & \textbf{CIS} & \textbf{CPS} & \textbf{CIS} & \textbf{CPS} & \textbf{Vote} \\
         \midrule
         \centering 2 & \multirow{5}{*}{\centering 47.13} & 48.75 & 51.00 & 49.14 & 52.35 & 47.44 \\
         \centering 3 & & 49.25 & 51.11 & 50.03 & 52.88 & 48.31 \\
         \centering 4 & & 49.69 & 51.96 & 51.02 & 54.17 & 48.29 \\
         \centering 5 & & \textbf{50.86} & \textbf{52.70} & 50.95 & 53.90 & 47.65 \\
         \centering 6 & & 50.35 & 52.40 & \textbf{51.70} & \textbf{54.64} & \textbf{48.34} \\
         \bottomrule
      \end{tabular}
   }
   \caption{Comparison of LLM-ACU scores on MultiNews dataset using different ensemble methods. The vote scores are model-invariant and apply to both models. Baseline indicates single sample performance without prompt ensemble. Best score for each model and aggregation agent is shown in \textbf{bold}.}
   \label{tab:acu_multinews}
\end{table}

\begin{table}[t!]
   \centering
   \adjustbox{width=0.5\textwidth}{
      \setlength{\tabcolsep}{7pt}  
      \begin{tabular}{c c cc cc c}
         \toprule
         \multicolumn{2}{c}{\textbf{LLM-ACU (OpenASP)}} & \multicolumn{2}{c}{\textbf{gpt-4o}} & \multicolumn{2}{c}{\textbf{gpt-4o-mini}} & \\
         \cmidrule(lr){3-4} \cmidrule(lr){5-6}
         \textbf{\# Samples} & \textbf{Baseline} & \textbf{CIS} & \textbf{CPS} & \textbf{CIS} & \textbf{CPS} & \textbf{Vote} \\
         \midrule
         \centering 2 & \multirow{5}{*}{\centering 42.35} & 43.05 & 44.16 & 44.36 & 46.07 & 43.86 \\
         \centering 3 & & 44.00 & 45.00 & 45.04 & 47.35 & 44.03 \\
         \centering 4 & & 43.64 & 45.51 & 45.05 & 47.55 & 44.47 \\
         \centering 5 & & 44.07 & \textbf{46.47} & 46.13 & \textbf{47.82} & 44.47 \\
         \centering 6 & & \textbf{44.66} & 46.30 & \textbf{46.35} & 47.46 & \textbf{45.00} \\
         \bottomrule
      \end{tabular}
   }
   \caption{Comparison of LLM-ACU scores on OpenASP dataset under same settings as Table~\ref{tab:acu_multinews}.}
   \label{tab:acu_openasp}
\end{table}

\begin{table*}[!htbp]
\centering
\resizebox{0.8\textwidth}{!}{%
\begin{tabular}{ccc cccc cccc}
\toprule
\textbf{Dataset} & \textbf{Model} & \centering \textbf{\# Samples} & \multicolumn{4}{c}{\textbf{CPS}} & \multicolumn{4}{c}{\textbf{CIS}} \\
\cline{4-11}
& & & \textbf{R1} & \textbf{R2} & \textbf{RL} & \textbf{RLsum} & \textbf{R1} & \textbf{R2} & \textbf{RL} & \textbf{RLsum} \\
\midrule
\multirow{11}{*}{\textbf{MultiNews}} & \textbf{Baseline} & \centering 1 & 36.32 & 10.57 & 18.58 & 19.24 & 36.32 & 10.57 & 18.58 & 19.24 \\
\cmidrule{2-11}
& \multirow{5}{*}{\textbf{gpt-4o}} & \centering 2 & 37.53 & 10.51 & 18.43 & 18.96 & 36.34 & 10.08 & 18.16 & 18.40 \\
& & \centering 3 & 37.75 & 10.56 & 18.69 & 19.22 & 36.90 & 10.22 & 18.35 & 18.63 \\
& & \centering 4 & 37.77 & 10.61 & 18.65 & 19.27 & 37.00 & 10.31 & 18.40 & \textbf{18.75} \\
& & \centering 5 & \textbf{38.22} & \textbf{10.88} & 18.84 & \textbf{19.61} & 37.06 & 10.35 & \textbf{18.43} & 18.72 \\
& & \centering 6 & 38.17 & 10.83 & \textbf{18.87} & 19.57 & \textbf{37.26} & \textbf{10.43} & 18.42 & 18.71 \\
\cmidrule{2-11}
& \multirow{5}{*}{\textbf{gpt-4o-mini}} & \centering 2 & 39.05 & 10.77 & 18.83 & 20.92 & 37.07 & 10.13 & 18.25 & 18.89 \\
& & \centering 3 & 39.24 & 10.87 & 18.88 & 21.15 & 37.49 & 10.14 & 18.39 & 19.29 \\
& & \centering 4 & 39.40 & 10.86 & 18.87 & 21.37 & 37.83 & 10.21 & 18.41 & 19.60 \\
& & \centering 5 & 39.44 & 10.92 & 18.89 & 21.45 & 38.09 & 10.41 & 18.49 & 19.86 \\
& & \centering 6 & \textbf{39.67} & \textbf{11.05} & \textbf{18.91} & \textbf{21.54} & \textbf{38.35} & \textbf{10.49} & \textbf{18.68} & \textbf{20.21} \\
\midrule
\multirow{11}{*}{\textbf{OpenASP}} & \textbf{Baseline} & \centering 1 & 32.47 & 7.89 & 15.77 & 17.11 & 32.47 & 7.89 & 15.77 & 17.11 \\
\cmidrule{2-11}
& \multirow{5}{*}{\textbf{gpt-4o}} & \centering 2 & 33.33 & 7.82 & 15.93 & 17.51 & 32.19 & 7.50 & 15.60 & 16.49 \\
& & \centering 3 & 33.37 & 7.88 & 15.91 & 17.54 & 32.38 & 7.43 & 15.62 & 16.69 \\
& & \centering 4 & 33.68 & 7.96 & 16.06 & 17.86 & 32.41 & 7.49 & 15.70 & 16.89 \\
& & \centering 5 & 33.72 & 8.06 & 16.01 & 17.89 & 32.64 & \textbf{7.66} & \textbf{15.68} & 16.92 \\
& & \centering 6 & \textbf{33.96} & \textbf{8.09} & \textbf{16.09} & \textbf{18.00} & \textbf{32.71} & 7.62 & 15.67 & \textbf{17.01} \\
\cmidrule{2-11}
& \multirow{5}{*}{\textbf{gpt-4o-mini}} & \centering 2 & 35.36 & 8.14 & 16.20 & 19.56 & 33.17 & 7.56 & 15.75 & 17.79 \\
& & \centering 3 & 35.79 & 8.31 & 16.23 & 19.83 & 33.90 & 7.65 & 15.91 & 18.39 \\
& & \centering 4 & 35.70 & 8.29 & 16.25 & 19.93 & 34.27 & 7.83 & 15.96 & 18.65 \\
& & \centering 5 & 35.94 & 8.31 & \textbf{16.37} & 20.07 & \textbf{34.55} & \textbf{7.83} & \textbf{16.01} & \textbf{18.94} \\
& & \centering 6 & \textbf{36.05} & \textbf{8.38} & 16.31 & \textbf{20.14} & 34.51 & 7.79 & 15.94 & 18.87 \\
\bottomrule
\end{tabular}%
}
\caption{Comparison of ROUGE scores on MultiNews and OpenASP datasets using different models and ensemble sizes. Best score for each dataset, model and aggregation method is shown in \textbf{bold}.}
\label{tab:rouge_combined}
\end{table*}
\subsection{Scaling Boundaries and Inverse Scaling}

\begin{figure}[htbp]  
    \centering
    \includegraphics[width=0.8\linewidth]{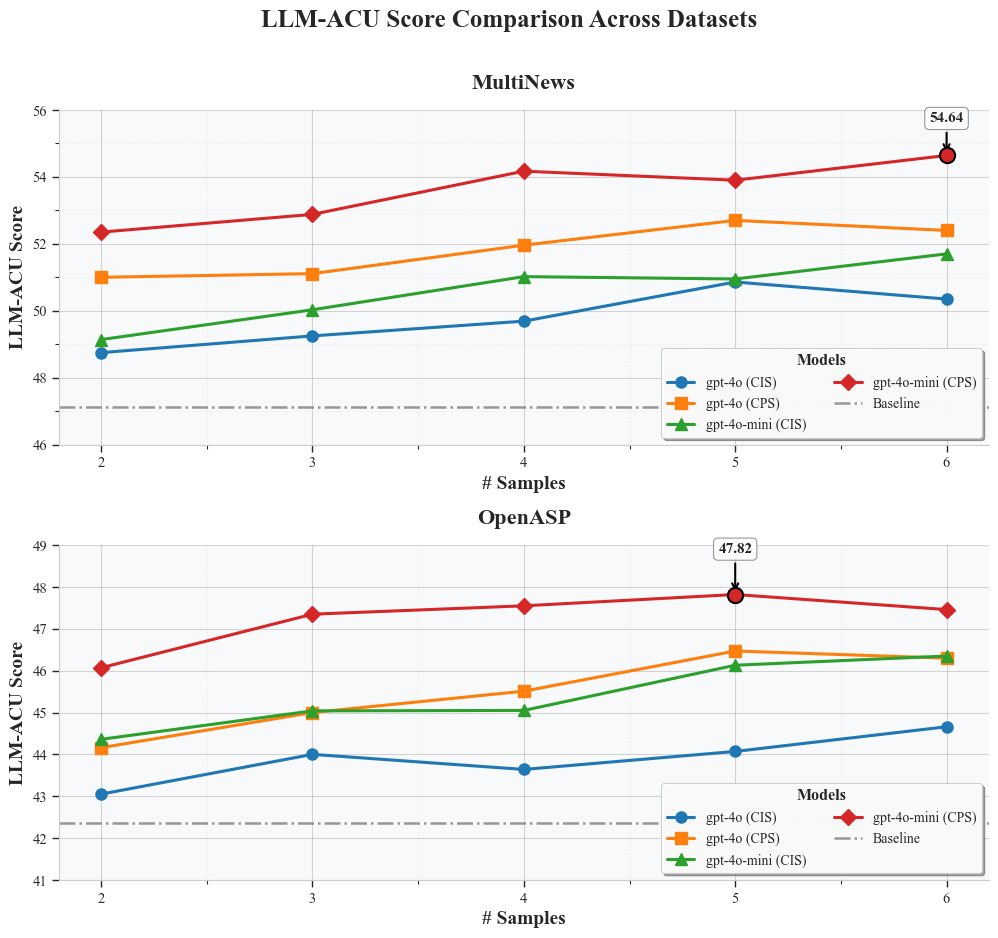}
    \caption{Scaling boundaries based on completeness (LLM-ACU) scores. Across both datasets, CPS consistently outperforms CIS. The smaller model (\texttt{gpt-4o-mini}) demonstrates greater performance gains from ensemble scaling, with improvements continuing at higher sample counts compared to the larger model (\texttt{gpt-4o}), which reaches its scaling plateau earlier.}
    \label{fig:acuscale}
\end{figure}
The scaling limitations manifest differently across ensemble methods. In terms of completeness, voting shows minimal improvement across all sample sizes, suggesting that simple selection-based ensemble may be insufficient for maintaining comprehensive information coverage. The impact of document context during ensemble emerges as a crucial factor. While CIS performs better than voting, it consistently achieves lower completeness scores than CPS, indicating that losing document context during ensemble creates a ceiling on information preservation. 

The scaling boundaries for CIS and CPS are clearly illustrated in Figure~\ref{fig:capscale} (preference score represented by $\text{CAP}_{max}$) and Figure~\ref{fig:acuscale} (completeness score represented by LLM-ACU). For preference scores, both datasets exhibit saturation points at approximately 5 samples, beyond which additional scaling yields diminishing returns. This inverse scaling phenomenon is particularly evident in MultiNews, where CPS performance peaks at 5 samples before declining at 6 samples, with the preference score nearly dropping to the same level as CIS. Completeness metrics follow a similar pattern, with \texttt{gpt-4o}'s scores using CPS plateauing around 5 samples, and \texttt{gpt-4o-mini} demonstrating comparable saturation behavior.

These observations suggest that excessive ensemble sizes may introduce noise rather than improvements, and that the choice of ensemble method significantly affects both quality and coverage outcomes. \textbf{This highlights the importance of identifying optimal scaling thresholds and maintaining document context throughout the ensemble process.}

\subsection{Scaling Effect across Model Sizes}
Our experiments with \texttt{gpt-4o} and \texttt{gpt-4o-mini} reveal interesting patterns in how model size interacts with scaling benefits.
In terms of completeness scores, \texttt{gpt-4o-mini} often achieves larger relative improvements compared to \texttt{gpt-4o} when scaled through prompt ensemble. This suggests that \textbf{prompt ensemble can partially compensate for model size limitations in terms of information capture.}

Regarding preference scores, the relationship between model size and performance is more nuanced. While \texttt{gpt-4o} generally outperforms \texttt{gpt-4o-mini} on MultiNews when using CPS, the smaller model achieves competitive results with CIS. More surprisingly, on OpenASP, \texttt{gpt-4o-mini} consistently outperforms its larger version across both CIS and CPS aggregators. This suggests that the \textbf{benefits of model scale are not uniform across different summarization tasks, and that scaling smaller models, when combined with appropriate scaling strategies, may sometimes be more effective.}
These findings challenge the assumption that larger models necessarily benefit more from inference-time scaling and emphasize the importance of considering both model size and ensemble size in optimization strategies.

\section{Conclusion}
In this work, we introduced the Multi$^2$ framework to scale MDS through prompt ensemble, showing that we can leverage computational resources at test time to produce more comprehensive and accurate summaries. Our evaluation metrics—CAP score and LLM-ACU score—also provide more reliable assessments by effectively mitigating positional bias in summary evaluation. Through systematic analysis, we identified specific scaling boundaries in summarization tasks, offering valuable insights into scaling summarization.
Our findings suggest two promising research directions: (1) incorporating test-time search algorithms to dynamically guide prompt ensemble optimization, and (2) extending our evaluation metrics to assess model performance in reasoning tasks. These directions highlight the potential of optimizing LLMs' inference-time behavior across applications where both factual accuracy and logical consistency are crucial.

\section*{Limitations}
Despite demonstrating that test-time scaling improves summarization quality, our work has several limitations.
First, we restricted our experimental scope to larger general-purpose commercial LLMs rather than including smaller open-source LLMs. This decision was guided by two considerations: (1) our primary objective was to validate the Multi$^2$ framework's general effectiveness rather than comprehensively benchmarking various LLMs' scaling capabilities, and (2) MDS tasks require robust context understanding typically found in general-purpose, market-proven models rather than smaller models with limited contextual processing ability.
Second, we did not conduct large-scale human evaluations to compare alignment between our metrics and previous ones. This decision reflects that the baseline metrics we sought to improve have already undergone comprehensive human evaluation and peer review, making additional human studies redundant for our specific research questions.

\bibliography{custom}

\begin{thebibliography}{68}
\providecommand{\natexlab}[1]{#1}

\bibitem[{Ackley et~al.(1985)Ackley, Hinton, and Sejnowski}]{ACKLEY1985147}
David~H. Ackley, Geoffrey~E. Hinton, and Terrence~J. Sejnowski. 1985.
\newblock \href {https://doi.org/10.1016/S0364-0213(85)80012-4} {A learning algorithm for boltzmann machines}.
\newblock \emph{Cognitive Science}, 9(1):147--169.

\bibitem[{Agarwal et~al.(2024)Agarwal, Singh, Zhang, Bohnet, Rosias, Chan, Zhang, Anand, Abbas, Nova, Co-Reyes, Chu, Behbahani, Faust, and Larochelle}]{agarwal2024manyshotincontextlearning}
Rishabh Agarwal, Avi Singh, Lei~M. Zhang, Bernd Bohnet, Luis Rosias, Stephanie Chan, Biao Zhang, Ankesh Anand, Zaheer Abbas, Azade Nova, John~D. Co-Reyes, Eric Chu, Feryal Behbahani, Aleksandra Faust, and Hugo Larochelle. 2024.
\newblock \href {https://arxiv.org/abs/2404.11018} {Many-shot in-context learning}.
\newblock \emph{Preprint}, arXiv:2404.11018.

\bibitem[{Allen-Zhu and Li(2024)}]{allenzhu2024physicslanguagemodels31}
Zeyuan Allen-Zhu and Yuanzhi Li. 2024.
\newblock \href {https://arxiv.org/abs/2309.14316} {Physics of language models: Part 3.1, knowledge storage and extraction}.
\newblock \emph{Preprint}, arXiv:2309.14316.

\bibitem[{Amar et~al.(2023)Amar, Schiff, Ernst, Shefer, Shapira, and Dagan}]{amar-etal-2023-openasp}
Shmuel Amar, Liat Schiff, Ori Ernst, Asi Shefer, Ori Shapira, and Ido Dagan. 2023.
\newblock \href {https://doi.org/10.18653/v1/2023.emnlp-main.121} {{O}pen{A}sp: A benchmark for multi-document open aspect-based summarization}.
\newblock In \emph{Proceedings of the 2023 Conference on Empirical Methods in Natural Language Processing}, pages 1967--1991, Singapore. Association for Computational Linguistics.

\bibitem[{Arora et~al.(2022)Arora, Narayan, Chen, Orr, Guha, Bhatia, Chami, Sala, and Ré}]{arora2022askanythingsimplestrategy}
Simran Arora, Avanika Narayan, Mayee~F. Chen, Laurel Orr, Neel Guha, Kush Bhatia, Ines Chami, Frederic Sala, and Christopher Ré. 2022.
\newblock \href {https://arxiv.org/abs/2210.02441} {Ask me anything: A simple strategy for prompting language models}.
\newblock \emph{Preprint}, arXiv:2210.02441.

\bibitem[{Belem et~al.(2024)Belem, Pezeskhpour, Iso, Maekawa, Bhutani, and Hruschka}]{belem2024singlemultillmshallucinate}
Catarina~G. Belem, Pouya Pezeskhpour, Hayate Iso, Seiji Maekawa, Nikita Bhutani, and Estevam Hruschka. 2024.
\newblock \href {https://arxiv.org/abs/2410.13961} {From single to multi: How llms hallucinate in multi-document summarization}.
\newblock \emph{Preprint}, arXiv:2410.13961.

\bibitem[{Bhandari et~al.(2020)Bhandari, Gour, Ashfaq, and Liu}]{bhandari-etal-2020-metrics}
Manik Bhandari, Pranav~Narayan Gour, Atabak Ashfaq, and Pengfei Liu. 2020.
\newblock \href {https://doi.org/10.18653/v1/2020.coling-main.501} {Metrics also disagree in the low scoring range: Revisiting summarization evaluation metrics}.
\newblock In \emph{Proceedings of the 28th International Conference on Computational Linguistics}, pages 5702--5711, Barcelona, Spain (Online). International Committee on Computational Linguistics.

\bibitem[{Brown et~al.(2024)Brown, Juravsky, Ehrlich, Clark, Le, Ré, and Mirhoseini}]{brown2024largelanguagemonkeysscaling}
Bradley Brown, Jordan Juravsky, Ryan Ehrlich, Ronald Clark, Quoc~V. Le, Christopher Ré, and Azalia Mirhoseini. 2024.
\newblock \href {https://arxiv.org/abs/2407.21787} {Large language monkeys: Scaling inference compute with repeated sampling}.
\newblock \emph{Preprint}, arXiv:2407.21787.

\bibitem[{Chen et~al.(2024)Chen, Davis, Hanin, Bailis, Stoica, Zaharia, and Zou}]{chen2024are}
Lingjiao Chen, Jared~Quincy Davis, Boris Hanin, Peter Bailis, Ion Stoica, Matei Zaharia, and James Zou. 2024.
\newblock \href {https://openreview.net/forum?id=m5106RRLgx} {Are more {LLM} calls all you need? towards the scaling properties of compound {AI} systems}.
\newblock In \emph{The Thirty-eighth Annual Conference on Neural Information Processing Systems}.

\bibitem[{Cobbe et~al.(2021)Cobbe, Kosaraju, Bavarian, Chen, Jun, Kaiser, Plappert, Tworek, Hilton, Nakano et~al.}]{cobbe2021training}
Karl Cobbe, Vineet Kosaraju, Mohammad Bavarian, Mark Chen, Heewoo Jun, Lukasz Kaiser, Matthias Plappert, Jerry Tworek, Jacob Hilton, Reiichiro Nakano, et~al. 2021.
\newblock Training verifiers to solve math word problems.
\newblock \emph{arXiv preprint arXiv:2110.14168}.

\bibitem[{Dubois et~al.(2024)Dubois, Liang, and Hashimoto}]{dubois2024lengthcontrolled}
Yann Dubois, Percy Liang, and Tatsunori Hashimoto. 2024.
\newblock \href {https://openreview.net/forum?id=CybBmzWBX0} {Length-controlled alpacaeval: A simple debiasing of automatic evaluators}.
\newblock In \emph{First Conference on Language Modeling}.

\bibitem[{Erkan and Radev(2004)}]{Erkan2004LexRankGL}
G{\"u}nes Erkan and Dragomir~R. Radev. 2004.
\newblock \href {https://api.semanticscholar.org/CorpusID:506350} {Lexrank: Graph-based lexical centrality as salience in text summarization}.
\newblock \emph{ArXiv}, abs/1109.2128.

\bibitem[{Fabbri et~al.(2019)Fabbri, Li, She, Li, and Radev}]{fabbri-etal-2019-multi}
Alexander Fabbri, Irene Li, Tianwei She, Suyi Li, and Dragomir Radev. 2019.
\newblock \href {https://doi.org/10.18653/v1/P19-1102} {Multi-news: A large-scale multi-document summarization dataset and abstractive hierarchical model}.
\newblock In \emph{Proceedings of the 57th Annual Meeting of the Association for Computational Linguistics}, pages 1074--1084, Florence, Italy. Association for Computational Linguistics.

\bibitem[{Gao et~al.(2022)Gao, Schulman, and Hilton}]{gao2022scalinglawsrewardmodel}
Leo Gao, John Schulman, and Jacob Hilton. 2022.
\newblock \href {https://arxiv.org/abs/2210.10760} {Scaling laws for reward model overoptimization}.
\newblock \emph{Preprint}, arXiv:2210.10760.

\bibitem[{Gerani et~al.(2014)Gerani, Mehdad, Carenini, Ng, and Nejat}]{gerani-etal-2014-abstractive}
Shima Gerani, Yashar Mehdad, Giuseppe Carenini, Raymond~T. Ng, and Bita Nejat. 2014.
\newblock \href {https://doi.org/10.3115/v1/D14-1168} {Abstractive summarization of product reviews using discourse structure}.
\newblock In \emph{Proceedings of the 2014 Conference on Empirical Methods in Natural Language Processing ({EMNLP})}, pages 1602--1613, Doha, Qatar. Association for Computational Linguistics.

\bibitem[{Giorgi et~al.(2022)Giorgi, Soldaini, Wang, Bader, Lo, Wang, and Cohan}]{Giorgi2022OpenDM}
John Giorgi, Luca Soldaini, Bo~Wang, Gary Bader, Kyle Lo, Lucy~Lu Wang, and Arman Cohan. 2022.
\newblock \href {https://api.semanticscholar.org/CorpusID:258865156} {Open domain multi-document summarization: A comprehensive study of model brittleness under retrieval}.
\newblock In \emph{Conference on Empirical Methods in Natural Language Processing}.

\bibitem[{Holtzman et~al.(2020)Holtzman, Buys, Du, Forbes, and Choi}]{DBLP:conf/iclr/HoltzmanBDFC20}
Ari Holtzman, Jan Buys, Li~Du, Maxwell Forbes, and Yejin Choi. 2020.
\newblock \href {https://openreview.net/forum?id=rygGQyrFvH} {The curious case of neural text degeneration}.
\newblock In \emph{8th International Conference on Learning Representations, {ICLR} 2020, Addis Ababa, Ethiopia, April 26-30, 2020}. OpenReview.net.

\bibitem[{Hu et~al.(2024)Hu, Song, Zhang, Xiao, Wang, Chen, Yuan, Lian, Ding, and Xiong}]{hu2024explaininglengthbiasllmbased}
Zhengyu Hu, Linxin Song, Jieyu Zhang, Zheyuan Xiao, Tianfu Wang, Zhengyu Chen, Nicholas~Jing Yuan, Jianxun Lian, Kaize Ding, and Hui Xiong. 2024.
\newblock \href {https://arxiv.org/abs/2407.01085} {Explaining length bias in llm-based preference evaluations}.
\newblock \emph{Preprint}, arXiv:2407.01085.

\bibitem[{Jin et~al.(2024)Jin, Xu, Zhang, Ling, Dong, Ouyang, Gao, Chang, and Sun}]{jin2024contranovo}
Zhi Jin, Sheng Xu, Xiang Zhang, Tianze Ling, Nanqing Dong, Wanli Ouyang, Zhiqiang Gao, Cheng Chang, and Siqi Sun. 2024.
\newblock Contranovo: A contrastive learning approach to enhance de novo peptide sequencing.
\newblock In \emph{Proceedings of the AAAI Conference on Artificial Intelligence}, volume~38, pages 144--152.

\bibitem[{Kojima et~al.(2022)Kojima, Gu, Reid, Matsuo, and Iwasawa}]{kojima2022large}
Takeshi Kojima, Shixiang~Shane Gu, Machel Reid, Yutaka Matsuo, and Yusuke Iwasawa. 2022.
\newblock \href {https://proceedings.neurips.cc/paper_files/paper/2022/hash/8bb0d291acd4acf06ef112099c16f326-Abstract-Conference.html} {Large language models are zero-shot reasoners}.
\newblock \emph{Advances in neural information processing systems}, 35:22199--22213.

\bibitem[{Li et~al.(2022)Li, Xiao, Xing, Wang, Murray, and Carenini}]{li2022humanguidedexploitationinterpretable}
Raymond Li, Wen Xiao, Linzi Xing, Lanjun Wang, Gabriel Murray, and Giuseppe Carenini. 2022.
\newblock \href {https://arxiv.org/abs/2112.05364} {Human guided exploitation of interpretable attention patterns in summarization and topic segmentation}.
\newblock \emph{Preprint}, arXiv:2112.05364.

\bibitem[{Li et~al.(2023{\natexlab{a}})Li, Lin, Zhang, Fu, Chen, Lou, and Chen}]{li-etal-2023-making}
Yifei Li, Zeqi Lin, Shizhuo Zhang, Qiang Fu, Bei Chen, Jian-Guang Lou, and Weizhu Chen. 2023{\natexlab{a}}.
\newblock \href {https://doi.org/10.18653/v1/2023.acl-long.291} {Making language models better reasoners with step-aware verifier}.
\newblock In \emph{Proceedings of the 61st Annual Meeting of the Association for Computational Linguistics (Volume 1: Long Papers)}, pages 5315--5333, Toronto, Canada. Association for Computational Linguistics.

\bibitem[{Li et~al.(2023{\natexlab{b}})Li, Lin, Zhang, Fu, Chen, Lou, and Chen}]{li2023makinglargelanguagemodels}
Yifei Li, Zeqi Lin, Shizhuo Zhang, Qiang Fu, Bei Chen, Jian-Guang Lou, and Weizhu Chen. 2023{\natexlab{b}}.
\newblock \href {https://arxiv.org/abs/2206.02336} {Making large language models better reasoners with step-aware verifier}.
\newblock \emph{Preprint}, arXiv:2206.02336.

\bibitem[{Lin(2004)}]{lin2004rouge}
Chin-Yew Lin. 2004.
\newblock Rouge: A package for automatic evaluation of summaries.
\newblock In \emph{Text summarization branches out}, pages 74--81.

\bibitem[{Lior et~al.(2024)Lior, Caciularu, Cattan, Levy, Shapira, and Stanovsky}]{lior2024seamstochasticbenchmarkmultidocument}
Gili Lior, Avi Caciularu, Arie Cattan, Shahar Levy, Ori Shapira, and Gabriel Stanovsky. 2024.
\newblock \href {https://arxiv.org/abs/2406.16086} {Seam: A stochastic benchmark for multi-document tasks}.
\newblock \emph{Preprint}, arXiv:2406.16086.

\bibitem[{Liu et~al.(2022)Liu, Zhang, and Mou}]{liu2022character}
Puyuan Liu, Xiang Zhang, and Lili Mou. 2022.
\newblock A character-level length-control algorithm for non-autoregressive sentence summarization.
\newblock \emph{Advances in Neural Information Processing Systems}, 35:29101--29112.

\bibitem[{Liu and Lapata(2019)}]{liu-lapata-2019-hierarchical}
Yang Liu and Mirella Lapata. 2019.
\newblock \href {https://doi.org/10.18653/v1/P19-1500} {Hierarchical transformers for multi-document summarization}.
\newblock In \emph{Proceedings of the 57th Annual Meeting of the Association for Computational Linguistics}, pages 5070--5081, Florence, Italy. Association for Computational Linguistics.

\bibitem[{Liu et~al.(2024{\natexlab{a}})Liu, Zhou, Guo, Shareghi, Vuli{\'c}, Korhonen, and Collier}]{liu2024aligning}
Yinhong Liu, Han Zhou, Zhijiang Guo, Ehsan Shareghi, Ivan Vuli{\'c}, Anna Korhonen, and Nigel Collier. 2024{\natexlab{a}}.
\newblock Aligning with human judgement: The role of pairwise preference in large language model evaluators.
\newblock \emph{arXiv preprint arXiv:2403.16950}.

\bibitem[{Liu et~al.(2023{\natexlab{a}})Liu, Fabbri, Liu, Zhao, Nan, Han, Han, Joty, Wu, Xiong, and Radev}]{liu-etal-2023-revisiting}
Yixin Liu, Alex Fabbri, Pengfei Liu, Yilun Zhao, Linyong Nan, Ruilin Han, Simeng Han, Shafiq Joty, Chien-Sheng Wu, Caiming Xiong, and Dragomir Radev. 2023{\natexlab{a}}.
\newblock \href {https://doi.org/10.18653/v1/2023.acl-long.228} {Revisiting the gold standard: Grounding summarization evaluation with robust human evaluation}.
\newblock In \emph{Proceedings of the 61st Annual Meeting of the Association for Computational Linguistics (Volume 1: Long Papers)}, pages 4140--4170, Toronto, Canada. Association for Computational Linguistics.

\bibitem[{Liu et~al.(2024{\natexlab{b}})Liu, Fabbri, Chen, Zhao, Han, Joty, Liu, Radev, Wu, and Cohan}]{liu-etal-2024-benchmarking}
Yixin Liu, Alexander Fabbri, Jiawen Chen, Yilun Zhao, Simeng Han, Shafiq Joty, Pengfei Liu, Dragomir Radev, Chien-Sheng Wu, and Arman Cohan. 2024{\natexlab{b}}.
\newblock \href {https://doi.org/10.18653/v1/2024.findings-naacl.280} {Benchmarking generation and evaluation capabilities of large language models for instruction controllable summarization}.
\newblock In \emph{Findings of the Association for Computational Linguistics: NAACL 2024}, pages 4481--4501, Mexico City, Mexico. Association for Computational Linguistics.

\bibitem[{Liu et~al.(2023{\natexlab{b}})Liu, Fabbri, Zhao, Liu, Joty, Wu, Xiong, and Radev}]{liu-etal-2023-towards-interpretable}
Yixin Liu, Alexander Fabbri, Yilun Zhao, Pengfei Liu, Shafiq Joty, Chien-Sheng Wu, Caiming Xiong, and Dragomir Radev. 2023{\natexlab{b}}.
\newblock \href {https://doi.org/10.18653/v1/2023.emnlp-main.1018} {Towards interpretable and efficient automatic reference-based summarization evaluation}.
\newblock In \emph{Proceedings of the 2023 Conference on Empirical Methods in Natural Language Processing}, pages 16360--16368, Singapore. Association for Computational Linguistics.

\bibitem[{Liu et~al.(2024{\natexlab{c}})Liu, Shi, Fabbri, Zhao, Wang, Wu, Joty, and Cohan}]{liu2024reifereevaluatinginstructionfollowingevaluation}
Yixin Liu, Kejian Shi, Alexander~R. Fabbri, Yilun Zhao, Peifeng Wang, Chien-Sheng Wu, Shafiq Joty, and Arman Cohan. 2024{\natexlab{c}}.
\newblock \href {https://arxiv.org/abs/2410.07069} {Reife: Re-evaluating instruction-following evaluation}.
\newblock \emph{Preprint}, arXiv:2410.07069.

\bibitem[{Madaan et~al.(2023)Madaan, Tandon, Gupta, Hallinan, Gao, Wiegreffe, Alon, Dziri, Prabhumoye, Yang, Gupta, Majumder, Hermann, Welleck, Yazdanbakhsh, and Clark}]{madaan2023selfrefine}
Aman Madaan, Niket Tandon, Prakhar Gupta, Skyler Hallinan, Luyu Gao, Sarah Wiegreffe, Uri Alon, Nouha Dziri, Shrimai Prabhumoye, Yiming Yang, Shashank Gupta, Bodhisattwa~Prasad Majumder, Katherine Hermann, Sean Welleck, Amir Yazdanbakhsh, and Peter Clark. 2023.
\newblock \href {https://openreview.net/forum?id=S37hOerQLB} {Self-refine: Iterative refinement with self-feedback}.
\newblock In \emph{Thirty-seventh Conference on Neural Information Processing Systems}.

\bibitem[{Mehdad et~al.(2014)Mehdad, Carenini, and Ng}]{mehdad-etal-2014-abstractive}
Yashar Mehdad, Giuseppe Carenini, and Raymond~T. Ng. 2014.
\newblock \href {https://doi.org/10.3115/v1/P14-1115} {Abstractive summarization of spoken and written conversations based on phrasal queries}.
\newblock In \emph{Proceedings of the 52nd Annual Meeting of the Association for Computational Linguistics (Volume 1: Long Papers)}, pages 1220--1230, Baltimore, Maryland. Association for Computational Linguistics.

\bibitem[{Muennighoff et~al.(2025)Muennighoff, Yang, Shi, Li, Fei-Fei, Hajishirzi, Zettlemoyer, Liang, Candès, and Hashimoto}]{muennighoff2025s1simpletesttimescaling}
Niklas Muennighoff, Zitong Yang, Weijia Shi, Xiang~Lisa Li, Li~Fei-Fei, Hannaneh Hajishirzi, Luke Zettlemoyer, Percy Liang, Emmanuel Candès, and Tatsunori Hashimoto. 2025.
\newblock \href {https://arxiv.org/abs/2501.19393} {s1: Simple test-time scaling}.
\newblock \emph{Preprint}, arXiv:2501.19393.

\bibitem[{OpenAI(2024)}]{learningtoreason2024}
OpenAI. 2024.
\newblock Learning to reason with llms.
\newblock \url{https://openai.com/index/learning-to-reason-with-llms/} [Accessed: 01/08/2025].

\bibitem[{Pu et~al.(2023)Pu, Gao, and Wan}]{pu2023summarizationalmostdead}
Xiao Pu, Mingqi Gao, and Xiaojun Wan. 2023.
\newblock \href {https://arxiv.org/abs/2309.09558} {Summarization is (almost) dead}.
\newblock \emph{Preprint}, arXiv:2309.09558.

\bibitem[{Shi et~al.(2024)Shi, Ma, Liang, Ma, and Vosoughi}]{shi2024judgingjudgessystematicstudy}
Lin Shi, Chiyu Ma, Wenhua Liang, Weicheng Ma, and Soroush Vosoughi. 2024.
\newblock \href {https://arxiv.org/abs/2406.07791} {Judging the judges: A systematic study of position bias in llm-as-a-judge}.
\newblock \emph{Preprint}, arXiv:2406.07791.

\bibitem[{Snell et~al.(2024)Snell, Lee, Xu, and Kumar}]{snell2024scalingllmtesttimecompute}
Charlie Snell, Jaehoon Lee, Kelvin Xu, and Aviral Kumar. 2024.
\newblock \href {https://arxiv.org/abs/2408.03314} {Scaling llm test-time compute optimally can be more effective than scaling model parameters}.
\newblock \emph{Preprint}, arXiv:2408.03314.

\bibitem[{Stroebl et~al.(2024)Stroebl, Kapoor, and Narayanan}]{stroebl2024inferencescalingflawslimits}
Benedikt Stroebl, Sayash Kapoor, and Arvind Narayanan. 2024.
\newblock \href {https://arxiv.org/abs/2411.17501} {Inference scaling flaws: The limits of llm resampling with imperfect verifiers}.
\newblock \emph{Preprint}, arXiv:2411.17501.

\bibitem[{Tian et~al.(2024)Tian, Peng, Song, Jin, Yu, Mi, and Yu}]{tian2024selfimprovementllmsimaginationsearching}
Ye~Tian, Baolin Peng, Linfeng Song, Lifeng Jin, Dian Yu, Haitao Mi, and Dong Yu. 2024.
\newblock \href {https://arxiv.org/abs/2404.12253} {Toward self-improvement of llms via imagination, searching, and criticizing}.
\newblock \emph{Preprint}, arXiv:2404.12253.

\bibitem[{Van~Veen et~al.(2024)Van~Veen, Van~Uden, Blankemeier, Delbrouck, Aali, Bluethgen, Pareek, Polacin, Collins, Ahuja, Langlotz, Hom, Gatidis, Pauly, and Chaudhari}]{vanveen2024clinical}
Dave Van~Veen, Cara Van~Uden, Louis Blankemeier, Jean-Benoit Delbrouck, Asad Aali, Christian Bluethgen, Anuj Pareek, Malgorzata Polacin, William Collins, Neera Ahuja, Curtis~P. Langlotz, Jason Hom, Sergios Gatidis, John Pauly, and Akshay~S. Chaudhari. 2024.
\newblock \href {https://doi.org/10.1038/s41591-024-02855-5} {Adapted large language models can outperform medical experts in clinical text summarization}.
\newblock \emph{Nature Medicine}.

\bibitem[{Wang et~al.(2024{\natexlab{a}})Wang, Deng, Lyu, Zeng, He, Yan, and An}]{wang2024q}
Chaojie Wang, Yanchen Deng, Zhiyi Lyu, Liang Zeng, Jujie He, Shuicheng Yan, and Bo~An. 2024{\natexlab{a}}.
\newblock Q*: Improving multi-step reasoning for llms with deliberative planning.
\newblock \emph{arXiv preprint arXiv:2406.14283}.

\bibitem[{Wang et~al.(2024{\natexlab{b}})Wang, Wang, Athiwaratkun, Zhang, and Zou}]{wang2024mixtureofagentsenhanceslargelanguage}
Junlin Wang, Jue Wang, Ben Athiwaratkun, Ce~Zhang, and James Zou. 2024{\natexlab{b}}.
\newblock \href {https://arxiv.org/abs/2406.04692} {Mixture-of-agents enhances large language model capabilities}.
\newblock \emph{Preprint}, arXiv:2406.04692.

\bibitem[{Wang et~al.(2024{\natexlab{c}})Wang, Li, Chen, Cai, Zhu, Lin, Cao, Kong, Liu, Liu, and Sui}]{wang-etal-2024-large-language-models-fair}
Peiyi Wang, Lei Li, Liang Chen, Zefan Cai, Dawei Zhu, Binghuai Lin, Yunbo Cao, Lingpeng Kong, Qi~Liu, Tianyu Liu, and Zhifang Sui. 2024{\natexlab{c}}.
\newblock \href {https://doi.org/10.18653/v1/2024.acl-long.511} {Large language models are not fair evaluators}.
\newblock In \emph{Proceedings of the 62nd Annual Meeting of the Association for Computational Linguistics (Volume 1: Long Papers)}, pages 9440--9450, Bangkok, Thailand. Association for Computational Linguistics.

\bibitem[{Wang et~al.(2023{\natexlab{a}})Wang, Li, Chen, Cai, Zhu, Lin, Cao, Liu, Liu, and Sui}]{wang2023large}
Peiyi Wang, Lei Li, Liang Chen, Zefan Cai, Dawei Zhu, Binghuai Lin, Yunbo Cao, Qi~Liu, Tianyu Liu, and Zhifang Sui. 2023{\natexlab{a}}.
\newblock Large language models are not fair evaluators.
\newblock \emph{arXiv preprint arXiv:2305.17926}.

\bibitem[{Wang et~al.(2023{\natexlab{b}})Wang, Wei, Schuurmans, Le, Chi, Narang, Chowdhery, and Zhou}]{wang2023selfconsistency}
Xuezhi Wang, Jason Wei, Dale Schuurmans, Quoc~V Le, Ed~H. Chi, Sharan Narang, Aakanksha Chowdhery, and Denny Zhou. 2023{\natexlab{b}}.
\newblock \href {https://openreview.net/forum?id=1PL1NIMMrw} {Self-consistency improves chain of thought reasoning in language models}.
\newblock In \emph{The Eleventh International Conference on Learning Representations}.

\bibitem[{Wei et~al.(2023)Wei, Wang, Schuurmans, Bosma, Ichter, Xia, Chi, Le, and Zhou}]{wei2023chainofthoughtpromptingelicitsreasoning}
Jason Wei, Xuezhi Wang, Dale Schuurmans, Maarten Bosma, Brian Ichter, Fei Xia, Ed~Chi, Quoc Le, and Denny Zhou. 2023.
\newblock \href {https://arxiv.org/abs/2201.11903} {Chain-of-thought prompting elicits reasoning in large language models}.
\newblock \emph{Preprint}, arXiv:2201.11903.

\bibitem[{Wu et~al.(2024)Wu, Sun, Li, Welleck, and Yang}]{wu2024scaling}
Yangzhen Wu, Zhiqing Sun, Shanda Li, Sean Welleck, and Yiming Yang. 2024.
\newblock \href {https://openreview.net/forum?id=j7DZWSc8qu} {Scaling inference computation: Compute-optimal inference for problem-solving with language models}.
\newblock In \emph{The 4th Workshop on Mathematical Reasoning and AI at NeurIPS'24}.

\bibitem[{Xiao et~al.(2024)Xiao, Xie, Carenini, and He}]{xiao-etal-2024-personalized}
Wen Xiao, Yujia Xie, Giuseppe Carenini, and Pengcheng He. 2024.
\newblock \href {https://aclanthology.org/2024.findings-eacl.39/} {Personalized abstractive summarization by tri-agent generation pipeline}.
\newblock In \emph{Findings of the Association for Computational Linguistics: EACL 2024}, pages 570--581, St. Julian{'}s, Malta. Association for Computational Linguistics.

\bibitem[{Yao et~al.(2023)Yao, Zhao, Yu, Du, Shafran, Narasimhan, and Cao}]{yao2023reactsynergizingreasoningacting}
Shunyu Yao, Jeffrey Zhao, Dian Yu, Nan Du, Izhak Shafran, Karthik Narasimhan, and Yuan Cao. 2023.
\newblock \href {https://arxiv.org/abs/2210.03629} {React: Synergizing reasoning and acting in language models}.
\newblock \emph{Preprint}, arXiv:2210.03629.

\bibitem[{Yin et~al.(2023)Yin, Kaddour, Zhang, Nie, Liu, Kong, and Liu}]{yin2023ttida}
Yuwei Yin, Jean Kaddour, Xiang Zhang, Yixin Nie, Zhenguang Liu, Lingpeng Kong, and Qi~Liu. 2023.
\newblock Ttida: Controllable generative data augmentation via text-to-text and text-to-image models.
\newblock \emph{arXiv preprint arXiv:2304.08821}.

\bibitem[{Yuan et~al.(2021)Yuan, Neubig, and Liu}]{Yuan2021BARTScoreEG}
Weizhe Yuan, Graham Neubig, and Pengfei Liu. 2021.
\newblock \href {https://api.semanticscholar.org/CorpusID:235593404} {Bartscore: Evaluating generated text as text generation}.
\newblock \emph{ArXiv}, abs/2106.11520.

\bibitem[{Zelikman et~al.(2024)Zelikman, Harik, Shao, Jayasiri, Haber, and Goodman}]{zelikman2024quietstarlanguagemodelsteach}
Eric Zelikman, Georges Harik, Yijia Shao, Varuna Jayasiri, Nick Haber, and Noah~D. Goodman. 2024.
\newblock \href {https://arxiv.org/abs/2403.09629} {Quiet-star: Language models can teach themselves to think before speaking}.
\newblock \emph{Preprint}, arXiv:2403.09629.

\bibitem[{Zhang et~al.(2024{\natexlab{a}})Zhang, Zhoubian, Hu, Yue, Dong, and Tang}]{zhang2024restmctsllmselftrainingprocess}
Dan Zhang, Sining Zhoubian, Ziniu Hu, Yisong Yue, Yuxiao Dong, and Jie Tang. 2024{\natexlab{a}}.
\newblock \href {https://arxiv.org/abs/2406.03816} {Rest-mcts*: Llm self-training via process reward guided tree search}.
\newblock \emph{Preprint}, arXiv:2406.03816.

\bibitem[{Zhang et~al.(2024{\natexlab{b}})Zhang, Wu, Lei, Che, Li, Xie, Huang, Zhang, Pavone, Li, Ouyang, and Zhou}]{zhang2024llamaberrypairwiseoptimizationo1like}
Di~Zhang, Jianbo Wu, Jingdi Lei, Tong Che, Jiatong Li, Tong Xie, Xiaoshui Huang, Shufei Zhang, Marco Pavone, Yuqiang Li, Wanli Ouyang, and Dongzhan Zhou. 2024{\natexlab{b}}.
\newblock \href {https://arxiv.org/abs/2410.02884} {Llama-berry: Pairwise optimization for o1-like olympiad-level mathematical reasoning}.
\newblock \emph{Preprint}, arXiv:2410.02884.

\bibitem[{Zhang et~al.(2020{\natexlab{a}})Zhang, Zhao, Saleh, and Liu}]{zhang2020pegasuspretrainingextractedgapsentences}
Jingqing Zhang, Yao Zhao, Mohammad Saleh, and Peter~J. Liu. 2020{\natexlab{a}}.
\newblock \href {https://arxiv.org/abs/1912.08777} {Pegasus: Pre-training with extracted gap-sentences for abstractive summarization}.
\newblock \emph{Preprint}, arXiv:1912.08777.

\bibitem[{Zhang et~al.(2020{\natexlab{b}})Zhang, Kishore, Wu, Weinberger, and Artzi}]{zhang2020bertscoreevaluatingtextgeneration}
Tianyi Zhang, Varsha Kishore, Felix Wu, Kilian~Q. Weinberger, and Yoav Artzi. 2020{\natexlab{b}}.
\newblock \href {https://arxiv.org/abs/1904.09675} {Bertscore: Evaluating text generation with bert}.
\newblock \emph{Preprint}, arXiv:1904.09675.

\bibitem[{Zhang et~al.(2024{\natexlab{c}})Zhang, Ladhak, Durmus, Liang, McKeown, and Hashimoto}]{zhang-etal-2024-benchmarking}
Tianyi Zhang, Faisal Ladhak, Esin Durmus, Percy Liang, Kathleen McKeown, and Tatsunori~B. Hashimoto. 2024{\natexlab{c}}.
\newblock \href {https://doi.org/10.1162/tacl_a_00632} {Benchmarking large language models for news summarization}.
\newblock \emph{Transactions of the Association for Computational Linguistics}, 12:39--57.

\bibitem[{Zhang et~al.(2024{\natexlab{d}})Zhang, Abdul-Mageed, and Lakshmanan}]{zhang2024autoregressive+}
Xiang Zhang, Muhammad Abdul-Mageed, and Laks~VS Lakshmanan. 2024{\natexlab{d}}.
\newblock Autoregressive+ chain of thought= recurrent: Recurrence's role in language models' computability and a revisit of recurrent transformer.
\newblock \emph{arXiv preprint arXiv:2409.09239}.

\bibitem[{Zhang et~al.(2024{\natexlab{e}})Zhang, Cao, and You}]{zhang2024counting}
Xiang Zhang, Juntai Cao, and Chenyu You. 2024{\natexlab{e}}.
\newblock Counting ability of large language models and impact of tokenization.
\newblock \emph{arXiv preprint arXiv:2410.19730}.

\bibitem[{Zhang and Ding(2024{\natexlab{a}})}]{zhang2024supervisedchainthought}
Xiang Zhang and Dujian Ding. 2024{\natexlab{a}}.
\newblock \href {https://arxiv.org/abs/2410.14198} {Supervised chain of thought}.
\newblock \emph{Preprint}, arXiv:2410.14198.

\bibitem[{Zhang and Ding(2024{\natexlab{b}})}]{zhang2024supervised}
Xiang Zhang and Dujian Ding. 2024{\natexlab{b}}.
\newblock Supervised chain of thought.
\newblock \emph{arXiv preprint arXiv:2410.14198}.

\bibitem[{Zhang et~al.(2023{\natexlab{a}})Zhang, Li, Hauer, Shi, and Kondrak}]{zhang-etal-2023-dont}
Xiang Zhang, Senyu Li, Bradley Hauer, Ning Shi, and Grzegorz Kondrak. 2023{\natexlab{a}}.
\newblock \href {https://doi.org/10.18653/v1/2023.emnlp-main.491} {Don`t trust {C}hat{GPT} when your question is not in {E}nglish: A study of multilingual abilities and types of {LLM}s}.
\newblock In \emph{Proceedings of the 2023 Conference on Empirical Methods in Natural Language Processing}, pages 7915--7927, Singapore. Association for Computational Linguistics.

\bibitem[{Zhang et~al.(2024{\natexlab{f}})Zhang, Li, Shi, Hauer, Wu, Kondrak, Abdul-Mageed, and Lakshmanan}]{zhang2024cross}
Xiang Zhang, Senyu Li, Ning Shi, Bradley Hauer, Zijun Wu, Grzegorz Kondrak, Muhammad Abdul-Mageed, and Laks~VS Lakshmanan. 2024{\natexlab{f}}.
\newblock Cross-modal consistency in multimodal large language models.
\newblock \emph{arXiv preprint arXiv:2411.09273}.

\bibitem[{Zhang et~al.(2023{\natexlab{b}})Zhang, Shi, Hauer, and Kondrak}]{zhang2023bridging}
Xiang Zhang, Ning Shi, Bradley Hauer, and Grzegorz Kondrak. 2023{\natexlab{b}}.
\newblock Bridging the gap between babelnet and hownet: Unsupervised sense alignment and sememe prediction.
\newblock In \emph{Proceedings of the 17th Conference of the European Chapter of the Association for Computational Linguistics}, pages 2789--2798.

\bibitem[{Zhao et~al.(2025)Zhao, Awasthi, and Gollapudi}]{zhao2025samplescrutinizescaleeffective}
Eric Zhao, Pranjal Awasthi, and Sreenivas Gollapudi. 2025.
\newblock \href {https://arxiv.org/abs/2502.01839} {Sample, scrutinize and scale: Effective inference-time search by scaling verification}.
\newblock \emph{Preprint}, arXiv:2502.01839.

\bibitem[{Zhuang et~al.(2024)Zhuang, Chen, Yu, Mitra, Bursztyn, Rossi, Sarkhel, and Zhang}]{zhuang2024toolchain}
Yuchen Zhuang, Xiang Chen, Tong Yu, Saayan Mitra, Victor Bursztyn, Ryan~A. Rossi, Somdeb Sarkhel, and Chao Zhang. 2024.
\newblock \href {https://openreview.net/forum?id=B6pQxqUcT8} {Toolchain*: Efficient action space navigation in large language models with a* search}.
\newblock In \emph{The Twelfth International Conference on Learning Representations}.

\end{thebibliography}

\appendix

\section{Related Work}
\subsection{Test-time scaling}
Test-time scaling strategies can be broadly classified into three categories: repeated sampling, deliberative approaches, and self-refinement. \textbf{Repeated sampling} leverages techniques like temperature sampling \citep{ACKLEY1985147}, top-$k$, and top-$p$ sampling \citep{DBLP:conf/iclr/HoltzmanBDFC20} to generate diverse outputs, which are then enhanced through aggregation strategies such as majority voting \citep{wang2023selfconsistency}, weighted majority voting \citep{li-etal-2023-making}, or best-of-$n$ selection \citep{cobbe2021training}. 
Recent work \citep{brown2024largelanguagemonkeysscaling,wu2024scaling,stroebl2024inferencescalingflawslimits,zhao2025samplescrutinizescaleeffective} demonstrates that repeated sampling can significantly expand LLM capabilities across various domains. \textbf{Deliberative approaches} incorporate structured reasoning through methods like chain-of-thought prompting \citep{wei2023chainofthoughtpromptingelicitsreasoning} and tree search. These approaches range from informed search methods \citep{zhuang2024toolchain,wang2024q} to Monte Carlo Tree Search (MCTS) variants \citep{tian2024selfimprovementllmsimaginationsearching,zhang2024llamaberrypairwiseoptimizationo1like,zhang2024restmctsllmselftrainingprocess}. A key characteristic of tree search methods is to use process reward models (PRMs) to guide the search trajectory during generation \citep{yao2023reactsynergizingreasoningacting,zelikman2024quietstarlanguagemodelsteach}. 
\textbf{Self-refinement}  \citep{madaan2023selfrefine} enables models to iteratively improve their responses through self-critique and editing. Additionally, all categories of test-time scaling methods can be enhanced through model ensembling \citep{wang2024mixtureofagentsenhanceslargelanguage,jin2024contranovo,chen2024are} to combine the strengths of multiple models to achieve better performance.

 Yet tree search methods often struggle with the high-dimensional search space created by multiple source documents, making it computationally intensive to explore meaningful trajectories. Self-refinement approaches, which rely on iterative improvements, may lead to information loss as they tend to focus on refining a single perspective rather than maintaining diverse viewpoints from multiple documents. 
 In our work, we adopt the repeated sampling approach to scale MDS at test time, using diverse prompts to generate multiple perspectives that are then consolidated through specialized aggregation methods.

\subsection{Multi Document Summarization and Evaluation}

Multi-document summarization (MDS) has evolved significantly from traditional methods \citep{Erkan2004LexRankGL,mehdad-etal-2014-abstractive,gerani-etal-2014-abstractive} to modern approaches powered by deep neural networks \citep{liu-lapata-2019-hierarchical, zhang2020pegasuspretrainingextractedgapsentences, Giorgi2022OpenDM,li2022humanguidedexploitationinterpretable}.
The advent of LLMs has boosted MDS capabilities even further, with models demonstrating impressive zero- and few-shot performance \citep{zhang-etal-2024-benchmarking}. Recent work to improve LLMs' summarization abilities has shifted the focus from models' architectural modifications to exploring various prompting strategies \citep{xiao-etal-2024-personalized, liu-etal-2024-benchmarking,yin2023ttida}. Despite these advances, MDS continues to face challenges including maintaining cross-document consistency, ensuring factual accuracy, and addressing content incompleteness where key information may be omitted \citep{belem2024singlemultillmshallucinate}. In this paper, we propose a test-time approach that addresses these challenges by generating summaries more aligned with user preferences.
Traditional evaluation metrics for summarization, such as ROUGE \citep{lin2004rouge}, only rely on lexical overlap with reference summaries. These metrics often fail to capture semantic similarity and summary quality adequately \citep{bhandari-etal-2020-metrics}. This limitation has led to the development of learned metrics that better align with human judgments \citep{Yuan2021BARTScoreEG,zhang2020bertscoreevaluatingtextgeneration}.
The emergence of LLMs has enabled even more sophisticated evaluation approaches. Recent work has explored using LLMs as evaluation agents \citep{liu-etal-2024-benchmarking, liu2024reifereevaluatinginstructionfollowingevaluation}, demonstrating their ability to assess multiple quality dimensions including coherence, faithfulness, and informativeness. However, these approaches face challenges such as positional bias and inconsistency across different model sizes \citep{wang-etal-2024-large-language-models-fair, shi2024judgingjudgessystematicstudy}. In this paper, we also try to address these limitations by proposing two metrics that remain consistent regardless of position or choice of evaluation model.

\section{Prompt Space Theory}
\label{app:theory}
In this section, we formalize the notion of the \emph{prompt space} and analyze its complexity in the context of Chain-of-Thought (CoT) reasoning. The prompt space, denoted as \(\mathcal{P}\), represents the set of all possible step templates that a language model (LM) may generate or be guided to generate during the reasoning process. Each template \(p \in \mathcal{P}\) is a discrete instruction that dictates how information is to be extracted from the latent representation \(h \in {R}^d\) and subsequently discretized into a sequence of tokens \(o = (o_1, o_2, \dots, o_k)\). In effect, the prompt space forms the interface between the continuous latent space and the discrete textual output~\cite{zhang2024autoregressive+}.

The latent vector \(h\) is assumed to encode \(m\) bits of information relevant to the task at hand. When the model follows a given prompt template \(p\), it extracts up to \(s\) bits of information per reasoning step. Thus, each template can be viewed as a function
\[
p: h \rightarrow o, \quad o \in \{0,1\}^s,
\]
where the mapping is constrained by the model’s capacity to “read out” a subset of the information encoded in \(h\). The total number of unique ways to extract \(s\) bits from \(m\) bits is given combinatorially by
\[
C(m,s) = \binom{m}{s} = \frac{m!}{s!(m-s)!}.
\]
This expression characterizes the \emph{prompt space complexity}, as it represents the number of potential step templates available to the model at each CoT step.

In practice, the prompt space is not uniformly sampled; instead, the LM employs learned heuristics to navigate this enormous space. That is, while the theoretical upper bound \(C(m,s)\) may be astronomically high, the effective search space is significantly reduced through task-specific training and, in many cases, human supervision. In an unsupervised setting, the model’s intrinsic biases might lead it to select suboptimal templates, thereby increasing the difficulty of navigating the subsequent \emph{answer space} \( \mathcal{S} \) – the space of all possible reasoning paths and final outputs.

More formally, let \(\phi\) denote the underlying computation that updates the hidden state:
\[
h_{t+1} = \phi(h_t, p),
\]
For brevity, we summarize the CoT process as follows: for \(t = 1, \dots, T\),
\[
o_t = p_t(h_{t-1}), \quad h_t = \phi(h_{t-1}, p_t).
\]
This compact notation encapsulates the iterative extraction of output tokens \(o_t\) and the recurrent update of the hidden state \(h_t\) via the chosen prompt \(p_t\).

Here, the selection of each \(p_t \in \mathcal{P}\) not only determines the immediate output \(o_t\) but also has a cascading effect on the evolution of the hidden state \(h_t\) and, consequently, the trajectory within the answer space \(\mathcal{S}\).

This intricate relationship between the prompt space and the answer space can be seen as a two-tier search problem: first, the model must identify a suitable template \(p\) from the high-dimensional prompt space \(\mathcal{P}\), and then it must effectively navigate the answer space \(\mathcal{S}\) defined by the recurrence \(h_t \rightarrow h_{t+1}\). Empirical evidence shows that even small deviations in the chosen template \(p\) can lead to exponentially larger errors in the final answer, underscoring the sensitivity of the overall reasoning process to prompt selection.

In summary, the prompt space theory emphasizes that the effectiveness of CoT reasoning hinges on the model’s ability to manage the combinatorial complexity inherent in extracting relevant information from its latent space. Supervised methods, which incorporate task-specific guidance, can significantly reduce the search complexity from the theoretical bound \(C(m,s)\) by constraining the model to a subset of high-quality prompts. This not only simplifies the navigation of the answer space but also enhances the overall reliability of the reasoning process. The theoretical insights presented in this work build directly upon the concepts introduced by \citet{zhang2024supervised}.

\section{Positional Bias in Automatic Evaluation}
\label{app:posbias}
In this section, we analyze the positional bias and consistency of two mainstream LLMs (\texttt{gpt-4o} and \texttt{claude-3.5-sonnet}). 

Tables~\ref{tab:modelpref1} and~\ref{tab:modelpref2} demonstrate a clear positional bias in both models' evaluations, though in opposing directions. \texttt{gpt-4o} shows a strong preference for summaries presented in the first position, with notably higher win ratios across both datasets. Conversely, \texttt{claude-3.5-sonnet} exhibits a preference for summaries in the second position, though this bias is relatively less pronounced in the MultiNews dataset.
This positional bias is further confirmed in Table~\ref{tab:modelconsist}, where the inconsistency ratios tell a similar story. The discrepancy percentages indicate that \texttt{claude-3.5-sonnet} generally achieves better consistency on MultiNews, though both models show comparable discrepancy rates on OpenASP.
While claude demonstrates marginally better consistency metrics overall, we opted to use \texttt{gpt-4o} in our final implementation due to practical considerations regarding speed and computational budget constraints. Since our evaluation framework incorporates both consistency and preference metrics, the choice between these models does not significantly impact the validity of our methodology or results.

These findings suggest that positional bias is still an inherent challenge in current language models when performing comparative evaluations, regardless of the specific model architecture or training approach. This observation underscores the importance of implementing appropriate debiasing strategies in evaluation frameworks.

\begin{table}[htbp]
\small
\centering
\begin{tabular}{llcc}
\toprule
\textbf{Model} & \textbf{Dataset} & \textbf{Sum1 Win} & \textbf{Sum2 Win} \\ \midrule
GPT    & MultiNews & 456 & 92 \\
Claude & MultiNews & 262 & 336 \\
GPT    & OpenASP   & 355 & 177 \\
Claude & OpenASP   & 186 & 401 \\
\bottomrule
\end{tabular}
\caption{Model Preference Analysis - Number of wins when comparing summaries in order \{Sum1, Sum2\}.}
\label{tab:modelpref1}
\end{table}

\begin{table}[ht]
\small
\centering
\begin{tabular}{llcc}
\toprule
\textbf{Model} & \textbf{Dataset} & \textbf{Sum2 Win} & \textbf{Sum1 Win} \\ \midrule
GPT    & MultiNews & 468 & 86 \\
Claude & MultiNews & 285 & 308 \\
GPT    & OpenASP   & 384 & 174 \\
Claude & OpenASP   & 188 & 396 \\
\bottomrule
\end{tabular}
\caption{Model Preference Analysis - Number of wins when comparing summaries in order \{Sum2, Sum1\}.}
\label{tab:modelpref2}
\end{table}

\begin{table}[htbp]
\small
\centering
\scalebox{0.95}{
\begin{tabular}{llcc}
\toprule
\textbf{Model/Dataset} & \textbf{Disc.(\%)} & \textbf{Pref Pos} & \textbf{Inc. Ratio} \\ \midrule
GPT/MultiNews    & 56.00\% & 1 & 333:3 \\
Claude/MultiNews & 16.67\% & 2 & 27:73 \\
GPT/OpenASP      & 30.03\% & 1 & 174:5 \\
Claude/OpenASP   & 34.72\% & 2 & 6:217 \\
\bottomrule
\end{tabular}
}
\caption{Model Consistency Analysis - Comparing discrepancy rates, positional bias, and inconsistency ratios between \texttt{gpt-4o} and \texttt{claude-3.5-sonnet}.}
\label{tab:modelconsist}
\end{table}

\section{Impact of Summary Length}
\label{app:lenimpact}
In this section, we investigate the relationship between summary quality and length. Tables~\ref{tab:lenimp_multinews} and~\ref{tab:lenimp_openasp} present CAP scores, ROUGELSum scores, and the lengths of generated summaries.

For capable models like \texttt{gpt-4o}, we observe that despite improvements in CAP and ROUGELsum scores, summary length remains relatively stable. Notably, the highest-quality summaries are not necessarily the longest ones, demonstrating that sophisticated models can effectively distill core ideas into concise text.

In contrast, for less capable models like \texttt{gpt-4o-mini}, preferred and more complete summaries consistently tend to be longer, with summary length increasing proportionally with the number of ensembled samples. This suggests that smaller models may require more text to adequately capture information compared to their larger counterparts.

\begin{table*}[htbp]
\centering
\begin{tabular}{cc|ccc|ccc}
\toprule
\multicolumn{2}{c}{\textbf{MultiNews}} & \multicolumn{3}{c}{\textbf{CPS}} & \multicolumn{3}{c}{\textbf{CIS}} \\
\cline{1-8}
\textbf{Model} & \centering\textbf{\# Samples} & \textbf{CAP} & \textbf{RLsum} & \textbf{Gen\_len} & \textbf{CAP} & \textbf{RLsum} & \textbf{Gen\_len} \\
\midrule
\multirow{5}{*}{\textbf{gpt-4o}} & \centering 2 & 0.82 & 18.96 & 155.53 & 0.69 & 18.40 & 138.82 \\
& \centering 3 & 0.79 & 19.22 & 158.26 & 0.73 & 18.63 & 145.06 \\
& \centering 4 & 0.82 & 19.27 & 161.86 & 0.68 & 18.75 & 147.26 \\
& \centering 5 & \textbf{0.85} & \textbf{19.61} & 163.14 & 0.71 & 18.72 & 147.60 \\
& \centering 6 & 0.81 & 19.57 & 163.58 & 0.79 & 18.71 & 158.25 \\
\midrule
\multirow{5}{*}{\textbf{gpt-4o-mini}} & \centering 2 & 0.80 & 20.92 & 184.98 & 0.61 & 18.89 & 150.58 \\
& \centering 3 & 0.72 & 21.15 & 190.76 & 0.54 & 19.29 & 159.40 \\
& \centering 4 & \textbf{0.80} & 21.37 & 196.85 & 0.60 & 19.60 & 165.36 \\
& \centering 5 & 0.78 & 21.45 & 191.18 & 0.60 & 19.86 & 170.03 \\
& \centering 6 & 0.77 & \textbf{21.54} & 201.07 & 0.60 & 20.21 & 172.44 \\
\bottomrule
\end{tabular}
\caption{CAP scores, ROUGELsum scores, and generation lengths on MultiNews dataset for different models and ensemble sizes. The highest CAP and ROUGELsum scores are marked in \textbf{bold}.}
\label{tab:lenimp_multinews}
\end{table*}

\begin{table*}[htbp]
\centering
\begin{tabular}{cc|ccc|ccc}
\toprule
\multicolumn{2}{c}{\textbf{OpenASP}} & \multicolumn{3}{c}{\textbf{CPS}} & \multicolumn{3}{c}{\textbf{CIS}} \\
\cline{1-8}
\textbf{Model} & \centering\textbf{\# Samples} & \textbf{CAP} & \textbf{RLsum} & \textbf{Gen\_len} & \textbf{CAP} & \textbf{RLsum} & \textbf{Gen\_len} \\
\midrule
\multirow{5}{*}{\textbf{gpt-4o}} & \centering 2 & 0.70 & 17.51 & 198.27 & 0.63 & 16.49 & 167.58 \\
& \centering 3 & 0.76 & 17.54 & 187.06 & 0.72 & 16.69 & 172.25 \\
& \centering 4 & 0.74 & 17.86 & 191.66 & 0.72 & 16.89 & 173.33 \\
& \centering 5 & 0.76 & 17.89 & 194.89 & 0.74 & 16.92 & 192.11 \\
& \centering 6 & \textbf{0.77} & \textbf{18.00} & 194.59 & 0.74 & 17.01 & 178.73 \\
\midrule
\multirow{5}{*}{\textbf{gpt-4o-mini}} & \centering 2 & 0.79 & 19.56 & 196.67 & 0.73 & 17.79 & 234.36 \\
& \centering 3 & 0.83 & 19.83 & 209.07 & 0.75 & 18.39 & 245.63 \\
& \centering 4 & 0.83 & 19.93 & 216.40 & 0.77 & 18.65 & 251.47 \\
& \centering 5 & \textbf{0.86} & 20.07 & 222.88 & 0.82 & 18.94 & 256.02 \\
& \centering 6 & 0.85 & \textbf{20.14} & 224.05 & 0.81 & 18.87 & 257.55 \\
\bottomrule
\end{tabular}
\caption{CAP scores, ROUGELsum scores, and generation lengths on OpenASP dataset for different models and ensemble sizes. The highest CAP and ROUGELsum scores are marked in \textbf{bold}.}
\label{tab:lenimp_openasp}
\end{table*}

Moreover, previous work \citep{hu2024explaininglengthbiasllmbased,dubois2024lengthcontrolled} reveals LLM evaluation mechanisms tend to favor long summaries. This raises an important question: ``do longer summaries actually contain more useful information?'' To investigate this, we study the relationship between generation length and summary quality using the general-purpose MDS dataset MultiNews.

The results in Table~\ref{tab:gen_cost} demonstrate how different configurations of our framework affect summary length and the associated computational costs.
 While the summary length increases substantially from baseline to our most comprehensive setting (from 129.4 to 201.17 words), the computational cost grows more slowly, suggesting efficient information packaging. The CPS aggregator consistently produces longer summaries than CIS, particularly with \texttt{gpt-4o-mini}, indicating its effectiveness in capturing diverse information from source documents without introducing excessive computational overhead.

 \begin{table}[htbp]
\centering
\begin{tabular}{l|cc}
\toprule
\textbf{Experiment} & \textbf{\# Words} & \textbf{Word/ACU} \\
\midrule
Baseline                  & 129.4 & 17.03 \\
\midrule
\texttt{gpt-4o}/CIS         & 147.61 & 18.42 \\
\texttt{gpt-4o}/CPS         & 163.15 & 19.51 \\
\texttt{gpt-4o-mini}/CIS    & 172.45 & 20.74 \\
\texttt{gpt-4o-mini}/CPS    & 201.17 & 22.63 \\
\bottomrule
\end{tabular}
\caption{Summary length and word cost per ACU across different model configurations on MultiNews dataset. Length shows the average number of words in generated summaries, while Cost measures the average number of words needed to capture each ACU.}
\label{tab:gen_cost}
\end{table}

\section{Human Evaluation}
\label{app:humaneval}
To evaluate the effectiveness of our proposed scaling approach, we conducted a human evaluation study involving three graduate students. The evaluators were presented with 10 samples, each containing summaries generated by our three ensemble methods (Voting, CIS, and CPS) using the optimal ensemble size of 5, as determined by our automatic evaluation. Each sample also included the corresponding baseline summary for comparison.
The evaluation was structured as a preference-based comparison between each ensemble method and the baseline. For each sample, evaluators were asked to indicate their preference between the ensemble-generated summary and the baseline summary, resulting in 30 comparisons per method (10 samples $\times$ 3 aggregation types).
The human evaluation results strongly support the effectiveness of our proposed methods, particularly the CPS approach. The voting-based ensemble was preferred over the baseline in 60\% of cases (18/30 comparisons). The CIS method demonstrated stronger performance, being preferred in 76.7\% of comparisons (23/30). Most notably, the CPS method achieved unanimous preference, being chosen over the baseline in all comparisons (29/30).
These results demonstrate a clear hierarchy among the ensemble methods, with CPS showing superior performance in human evaluation. The strong preference for CPS (96.7\%) aligns with our automatic evaluation findings, confirming that the method produces summaries that are not only technically sound but also qualitatively superior from a human perspective. The significant improvement over both the baseline and other ensemble methods suggests that CPS effectively captures and maintains important aspects of text summarization that human readers value.

\section{Prompts}
\subsection{Summarization Prompts}
\label{app:summprompt}
In Tables~\ref{tab:promptbankmultinews1} and~\ref{tab:promptbankmultinews2}, we present the prompt bank used for the MultiNews dataset. Similarly, Tables~\ref{tab:promptbankopenasp1} and~\ref{tab:promptbankopenasp2} contain the prompt bank for the OpenASP dataset. These prompts were adapted and modified from the work of \citet{lior2024seamstochasticbenchmarkmultidocument}. We utilized the same few-shot examples as provided in their benchmark.

\subsection{Ensemble Prompts}
\label{app:ensembleprompt}
We present our summary ensemble prompts for general purpose MDS (for datasets like MultiNews) in Table~\ref{tab:ensemblegeneral}, and for aspect- (or query-) based MDS (for datasets like OpenASP) in Table~\ref{tab:ensembleaspect}.

\newpage
\definecolor{tableheader}{RGB}{52, 73, 94}
\definecolor{tablegray}{RGB}{245, 246, 250}

\begin{table*}[htbp]
    \centering
    \renewcommand{\arraystretch}{1.3}
    \begin{tabularx}{\textwidth}{|>{\columncolor{tablegray}}c|>{\columncolor{white}}X|}
        \hline
        \rowcolor{tableheader}
        \textbf{\textcolor{white}{No.}} & \textbf{\textcolor{white}{Prompt}} \\
        \hline
        1 & In this task, you are presented with multiple news articles about related topics. Your job is to generate a summary that integrates information from the provided articles. Your summary should be short and concise, that includes content only from the provided articles, avoiding any external data sources. \\
        \hline
        2 & Please provide a brief summary by synthesizing only the key points from the articles provided. Focus on the main arguments and conclusions without incorporating any information from outside these texts. Keep your summary concise and directly related to the content of the documents. \\
        \hline
        3 & Generate a concise summary using only the information from the provided articles. Your summary should distill the most essential information, capturing the core insights without adding any external content. Aim for brevity and clarity in your summarization. \\
        \hline
        4 & Please sift through the provided articles and distill their essence into a sharp, concise summary. Focus solely on the facts and key points within these texts, avoiding any embellishment or reference to external information. Your summary should read like a bullet-point list of the most critical insights. \\
        \hline
        5 & You are presented with multiple news articles about related topics. Summarize the contents in a way that captures the key information in a narrative form, but strictly using the details mentioned in the provided documents. Keep it engaging yet brief. \\
        \hline
        6 & Imagine you're preparing a brief for a decision-maker who has limited time. Summarize the provided documents by extracting only the most essential information. Present this in a clear, straightforward manner, focusing on the key facts and figures. \\
        \hline
        7 & Using only the details from the articles I've given you, craft a summary that distills the most important information. Avoid any interpretations or external data, and keep your summary short and direct. Emphasize the main arguments, data points, and conclusions. \\
        \hline
        8 & Operate as an information synthesizer: Draw the essence from multiple articles, focusing solely on the information contained within them. Your summary should be a tight, focused digest of the articles, free from any influence of external data. \\
        \hline
        9 & Scan through the provided articles and compile a summary that highlights only the most significant facts and figures, ensuring the exclusion of all external references. Aim for clarity and brevity. \\
        \hline
        10 & Operate as an academic summarizer: Imagine you are creating a summary for an academic review. Extract and emphasize the most pertinent information, ensuring your summary remains true to the original texts and free of external content. \\
        \hline
    \end{tabularx}
    \caption{Summarization Prompt Bank for MultiNews Dataset (Part 1)}
    \label{tab:promptbankmultinews1}
\end{table*}

\begin{table*}[htbp]
    \centering
    \renewcommand{\arraystretch}{1.3}
    \begin{tabularx}{\textwidth}{|>{\columncolor{tablegray}}c|>{\columncolor{white}}X|}
        \hline
        \rowcolor{tableheader}
        \textbf{\textcolor{white}{No.}} & \textbf{\textcolor{white}{Prompt}} \\
        \hline
        11 & Condense the provided information into a compact summary that emphasizes the main points and crucial data from the documents. Exclude any external information to maintain the integrity of the sources. \\
        \hline
        12 & From the provided articles, pull out the core messages and data points. Shape these into a brief, clear summary that directly reflects the content of the documents without any external additions. \\
        \hline
        13 & Compile a concise summary from the news articles given, focusing only on the information contained within. Your summary should integrate the main points without adding any outside information. \\
        \hline
        14 & Create a succinct summary by focusing exclusively on the details provided in the articles. Avoid using any external sources and ensure the summary remains clear and to the point. \\
        \hline
        15 & Produce a brief summary that distills the essential facts from the provided articles. Keep your summary strictly to the content presented in the documents, avoiding external influences. \\
        \hline
        16 & Develop a concise summary using only the information from the articles provided. Emphasize the main points and conclusions while avoiding the inclusion of any external data. \\
        \hline
        17 & Prepare a short, integrated summary by synthesizing key points from the given news articles. Ensure that no external content is included and that the summary is clear and direct. \\
        \hline
        18 & Your task is to distill the primary information from the provided articles into a concise summary. Make sure to exclude any external sources and focus strictly on the given texts. \\
        \hline
        19 & Summarize the provided articles by extracting only the key information and conclusions. Your summary should be brief and must not incorporate any external data. \\
        \hline
        20 & Generate a clear and brief summary using just the information from the provided articles. Focus on distilling the essential points and data without referencing external content. \\
        \hline
    \end{tabularx}
    \caption{Summarization Prompt Bank for MultiNews Dataset (Part 2)}
    \label{tab:promptbankmultinews2}
\end{table*}

\begin{table*}[htbp]
    \centering
    \renewcommand{\arraystretch}{1.3}
    \begin{tabularx}{\textwidth}{|>{\columncolor{tablegray}}c|>{\columncolor{white}}X|}
        \hline
        \rowcolor{tableheader}
        \textbf{\textcolor{white}{No.}} & \textbf{\textcolor{white}{Prompt}} \\
        \hline
        1 & In this task you are required to generate an aspect-based summary of a set of documents related the same topic. Please write a short, concise aspect-based summary, only summarize content from the above documents, avoiding any external data sources. \\
        \hline
        2 & Your goal is to create a short, concise aspect-based summary of the given documents. Summarize the key points accurately, using only the information from these documents and excluding any external sources. \\
        \hline
        3 & Produce a brief, aspect-based summary of the collection of documents on the same topic. Ensure your summary is concise and derived only from the provided documents, avoiding any external data sources. \\
        \hline
        4 & Your task is to generate a detailed yet concise aspect-based summary from a collection of documents that focus on the same topic. Begin by thoroughly examining each document to understand the main aspects and themes. Then, synthesize this information into a coherent summary that highlights the significant points. \\
        \hline
        5 & Given a set of documents related to a specific topic, generate a short, concise aspect-based summary. Ensure that the summary is based solely on the content of the documents provided. \\
        \hline
        6 & You will receive several documents on the same topic. Your task is to write a brief aspect-based summary, using only the information from the provided documents and excluding any external sources. \\
        \hline
        7 & You are tasked with generating an aspect-based summary of several documents. Summarize the content briefly and accurately, using only the information from the documents give. \\
        \hline
        8 & In this task, you are required to create an aspect-based summary of a set of documents all related to the same topic. Carefully read through each document and identify the key aspects discussed. Summarize these aspects in a concise manner, ensuring that your summary captures the essential points. \\
        \hline
        9 & You are tasked with producing an aspect-based summary for a series of documents related to the same topic. Start by analyzing each document to identify the critical aspects covered. Your goal is to condense this information into a clear and concise summary. \\
        \hline
        10 & Generate a concise aspect-based summary of the given documents. Focus on summarizing the content based solely on the information from these documents, avoiding any external sources. \\
        \hline
    \end{tabularx}
    \caption{Summarization Prompt Bank for OpenASP Dataset (Part 1)}
    \label{tab:promptbankopenasp1}
\end{table*}

\begin{table*}[htbp]
    \centering
    \renewcommand{\arraystretch}{1.3}
    \begin{tabularx}{\textwidth}{|>{\columncolor{tablegray}}c|>{\columncolor{white}}X|}
        \hline
        \rowcolor{tableheader}
        \textbf{\textcolor{white}{No.}} & \textbf{\textcolor{white}{Prompt}} \\
        \hline
        11 & Create a concise aspect-based summary for the provided set of documents. Focus on the main aspects and themes discussed in these documents, ensuring that your summary is based entirely on the content of the provided documents. \\
        \hline
        12 & Produce a short and precise aspect-based summary of the given documents. Identify the key aspects discussed in these documents and synthesize a concise summary based solely on the provided content. \\
        \hline
        13 & You will receive a collection of documents focused on the same topic. Your task is to create an aspect-based summary that highlights the key aspects discussed in these documents. Ensure your summary is brief and does not include any external information. \\
        \hline
        14 & You are provided with multiple documents related to a single topic. Your task is to generate an aspect-based summary that captures the main aspects discussed in these documents. Ensure your summary is concise and solely based on the provided texts. \\
        \hline
        15 & You are tasked with generating an aspect-based summary of several documents on the same topic. Carefully review each document, identify the main aspects, and write a brief summary that captures these aspects using only the provided documents. \\
        \hline
        16 & Your role is to create an educational summary for students using a collection of documents on the same topic. Focus on the main aspects that would help students understand the core concepts discussed in the documents. \\
        \hline
        17 & Imagine you are preparing a briefing for a busy executive who needs to understand the key aspects of several documents quickly. Summarize the most important points from these documents in a concise manner. \\
        \hline
        18 & As an advanced AI tasked with summarizing documents, your goal is to generate an aspect-based summary. Think of yourself as a summarization expert, extracting the most critical aspects from the documents provided. \\
        \hline
        19 & Imagine you are a journalist tasked with writing a summary article based on a series of documents related to a single topic. Identify the key aspects discussed in these documents and compose a brief, coherent summary. \\
        \hline
        20 & Your task is to act as a knowledge distiller, creating a concise aspect-based summary from a series of documents on the same topic. Focus on identifying and summarizing the critical aspects discussed in these documents. \\
        \hline
        21 & You are an AI assistant tasked with providing a summary for a set of documents related to a specific topic. Focus on the key aspects and themes discussed in these documents. Create a summary that captures these aspects in a concise manner, ensuring that your summary is based solely on the provided documents and excludes any external information.\\
        \hline
    \end{tabularx}
    \caption{Summarization Prompt Bank for OpenASP Dataset (Part 2)}
    \label{tab:promptbankopenasp2}
\end{table*}
\newpage

\definecolor{headerblue}{RGB}{47, 72, 88}
\definecolor{lightgray}{RGB}{249, 250, 251}
\definecolor{darktext}{RGB}{44, 62, 80}

\begin{table*}[htbp]
    \centering
    \renewcommand{\arraystretch}{1.4}
    \begin{tabularx}{\textwidth}{|>{\columncolor{lightgray}\raggedright\arraybackslash}p{0.18\textwidth}|>{\columncolor{white}}X|}
        \hline
        \rowcolor{headerblue}
        \textbf{\textcolor{white}{Ensemble Type}} & \textbf{\textcolor{white}{Content}} \\
        \hline
        
        \textbf{Vote} & 
        Provide your explanation, then select the best summary of the given documents based on clarity, accuracy, conciseness, and completeness.
        
        \vspace{1mm}
        Documents: \{doc\}
        \vspace{2mm}
        
        Summary 1: \{sum1\}
        \vspace{2mm}
        
        Summary 2: \{sum2\}
        \vspace{2mm}
        
        ...
        \vspace{2mm}
        
        Explanation: ``Your explanation here''
        \vspace{2mm}
        
        Decision: [1-5] \\
        \hline
        
        \textbf{CIS} & 
        Take all provided summaries into account and generate a better, cohesive summary. Combine and refine the content from the summaries to ensure clarity, accuracy, conciseness, and completeness. Provide the final summary directly.
        
        \vspace{1mm}
        Summary 1: \{sum1\}
        \vspace{2mm}
        
        Summary 2: \{sum2\}
        \vspace{2mm}
        
        ...
        \vspace{2mm}
        
        Final revised summary: \\
        \hline
        
        \textbf{CPS} & 
        Take all provided summaries into account and generate a better, cohesive summary of the given documents. Combine and refine the content from the summaries to ensure clarity, accuracy, conciseness, and completeness. Provide the final summary directly.

        \vspace{1mm}
        Documents: \{doc\}
        \vspace{2mm}
        
        Summary 1: \{sum1\}
        \vspace{2mm}
        
        Summary 2: \{sum2\}
        \vspace{2mm}
        
        ...
        \vspace{2mm}
        
        Final revised summary: \\
        \hline
    \end{tabularx}
    \caption{Ensemble Prompts for General MDS}
    \label{tab:ensemblegeneral}
\end{table*}

\begin{table*}[htbp]
    \centering
    \renewcommand{\arraystretch}{1.4}
    \begin{tabularx}{\textwidth}{|>{\columncolor{lightgray}\raggedright\arraybackslash}p{0.18\textwidth}|>{\columncolor{white}}X|}
        \hline
        \rowcolor{headerblue}
        \textbf{\textcolor{white}{Ensemble Type}} & \textbf{\textcolor{white}{Content}} \\
        \hline
        
        \textbf{Vote} & 
        Provide your explanation, then select the best summary of the given documents based on clarity, accuracy, conciseness, and completeness, focusing on the specified aspect.
        \vspace{2mm}
        
        Example Response:
        \vspace{1mm}
        
        Explanation: ``Your explanation here''
        \vspace{1mm}
        
        Decision: 1 (or 2 or 3 or 4 or 5)
        \vspace{2mm}
        
        Aspect: \{query\}
        \vspace{2mm}
        
        Documents: \{doc\}
        \vspace{2mm}
        
        Summary 1: \{sum1\}
        \vspace{2mm}
        
        Summary 2: \{sum2\}
        \vspace{2mm}
        
        ...
        \vspace{2mm}
        
        Response: \\
        \hline
        
        \textbf{CIS} & 
        Take all provided summaries into account and generate a better, cohesive summary, focusing on the specified aspect. Combine and refine the content from the summaries to ensure clarity, accuracy, conciseness, and completeness. Provide the final summary directly.
        \vspace{2mm}
        
        Aspect: \{query\}
        \vspace{2mm}
        
        Summary 1: \{sum1\}
        \vspace{2mm}
        
        Summary 2: \{sum2\}
        \vspace{2mm}
        
        ...
        \vspace{2mm}
        
        Final revised summary: \\
        \hline
        
        \textbf{CPS} & 
        Take all provided summaries into account and generate a better, cohesive summary of the given documents, focusing on the specified aspect. Combine and refine the content from the summaries to ensure clarity, accuracy, conciseness, and completeness. Provide the final summary directly.
        \vspace{2mm}
        
        Aspect: \{query\}
        \vspace{2mm}
        
        Documents: \{doc\}
        \vspace{2mm}
        
        Summary 1: \{sum1\}
        \vspace{2mm}
        
        Summary 2: \{sum2\}
        \vspace{2mm}
        
        ...
        \vspace{2mm}
        
        Final revised summary: \\
        \hline
    \end{tabularx}
    \caption{Ensemble Prompts for Aspect-based MDS}
    \label{tab:ensembleaspect}
\end{table*}

\end{document}